%% file: main.tex
\documentclass[10pt,twocolumn,letterpaper]{article}

\usepackage[pagenumbers]{cvpr} 

\input{preamble}

\definecolor{cvprblue}{rgb}{0.21,0.49,0.74}
\usepackage[pagebackref,breaklinks,colorlinks,allcolors=cvprblue]{hyperref}
\usepackage{algorithm}
\usepackage{algpseudocode}
\algrenewcommand\algorithmicrequire{\textbf{Input:}}
\algrenewcommand\algorithmicensure{\textbf{Output:}}
\definecolor{lightblue}{HTML}{3c78d8}
\definecolor{lightgreen}{HTML}{38761d}
\definecolor{lightpurple}{rgb}{0.9,0.85,1}

\def\paperID{11559} 
\def\confName{CVPR}
\def\confYear{2026}

\title{VLA Knows Its Limits: Adaptive Execution Horizons for Robot Policies}

\author{\textbf{Haoxuan Wang}$^{1,2}$,
\textbf{Gengyu Zhang}$^{1,2}$, \\
\textbf{Yan Yan}$^{1}$,
\textbf{Ramana Rao Kompella}$^{2}$,
\textbf{Gaowen Liu}$^{2,\dagger}$ \\
$^{1}$University of Illinois Chicago \quad
$^{2}$Cisco Research
}

\begin{document}
\maketitle
\input{sec/0_abstract}    
\input{sec/1_intro}
\input{sec/2_related}
\input{sec/3_method}
\input{sec/4_experiment}
\input{sec/5_conclusion}

{
    \small
    \bibliographystyle{ieeenat_fullname}
    \bibliography{main}
}

\input{sec/X_suppl}

\end{document}

%% file: preamble.tex
\usepackage{multirow}
\usepackage{adjustbox} 
\usepackage{pifont}
\usepackage{enumitem}
\let\oldding\ding
\renewcommand{\ding}[2][1]{\scalebox{#1}{\oldding{#2}}}
\usepackage{amsthm}  
\usepackage{amsmath} 
\newtheorem{proposition}{Proposition}
\newenvironment{prop}[1][]
  {\begin{proposition}[\textit{#1}]}
  {\end{proposition}}



%% file: sec/0_abstract.tex
\begin{abstract}
Action chunking has recently emerged as a standard practice in flow-based Vision-Language-Action (VLA) models. 
However, the effect and choice of the execution horizon—the number of actions to be executed from each predicted chunk—remains underexplored. 
In this work, we first show that varying the execution horizon leads to substantial performance deviations, with performance initially improving and then declining as the horizon increases.
To uncover the reasons, we analyze the cross- and self-attention weights in flow-based VLAs and reveal two key phenomena: 
(i) intra-chunk actions attend invariantly to vision–language tokens, limiting adaptability to environmental changes; and 
(ii) the initial and terminal action tokens serve as stable anchors, forming latent centers around which intermediate actions are organized. 
Motivated by these insights, we interpret action self-attention weights as a proxy for the model's predictive limit and propose \textbf{AutoHorizon}, the first test-time method that dynamically estimates the execution horizon for each predicted action chunk to adapt to changing perceptual conditions.
Across simulated and real-world robotic manipulation tasks, AutoHorizon is performant, incurs negligible computational overhead, and generalizes across diverse tasks and flow-based models. Video demos are available at this \href{https://hatchetproject.github.io/autohorizon/}{project page}.
\end{abstract}

%% file: sec/1_intro.tex
\section{Introduction}
\label{sec:intro}

As a key milestone toward artificial general intelligence, embodied AI seeks to endow agents with the ability to perceive, reason, and act within the physical world~\cite{vla_survey1,vla_survey2}. Recent advances in machine-learning–based control have enabled these agents to directly interact with real-world environments through learned representations~\cite{dp,dp3,act}. Within this paradigm, Vision-Language-Action (VLA) models~\cite{openvla,pi0,pi05,smolvla,3dvla,hirobot,uva} have emerged as a promising direction for their ability to ground visual perception and linguistic instruction into executable actions, demonstrating strong performance across diverse language-conditioned robot manipulation tasks. 

\begin{figure}[t]
    \centering
    \includegraphics[width=1.0\linewidth, trim=10 20 10 10, clip]{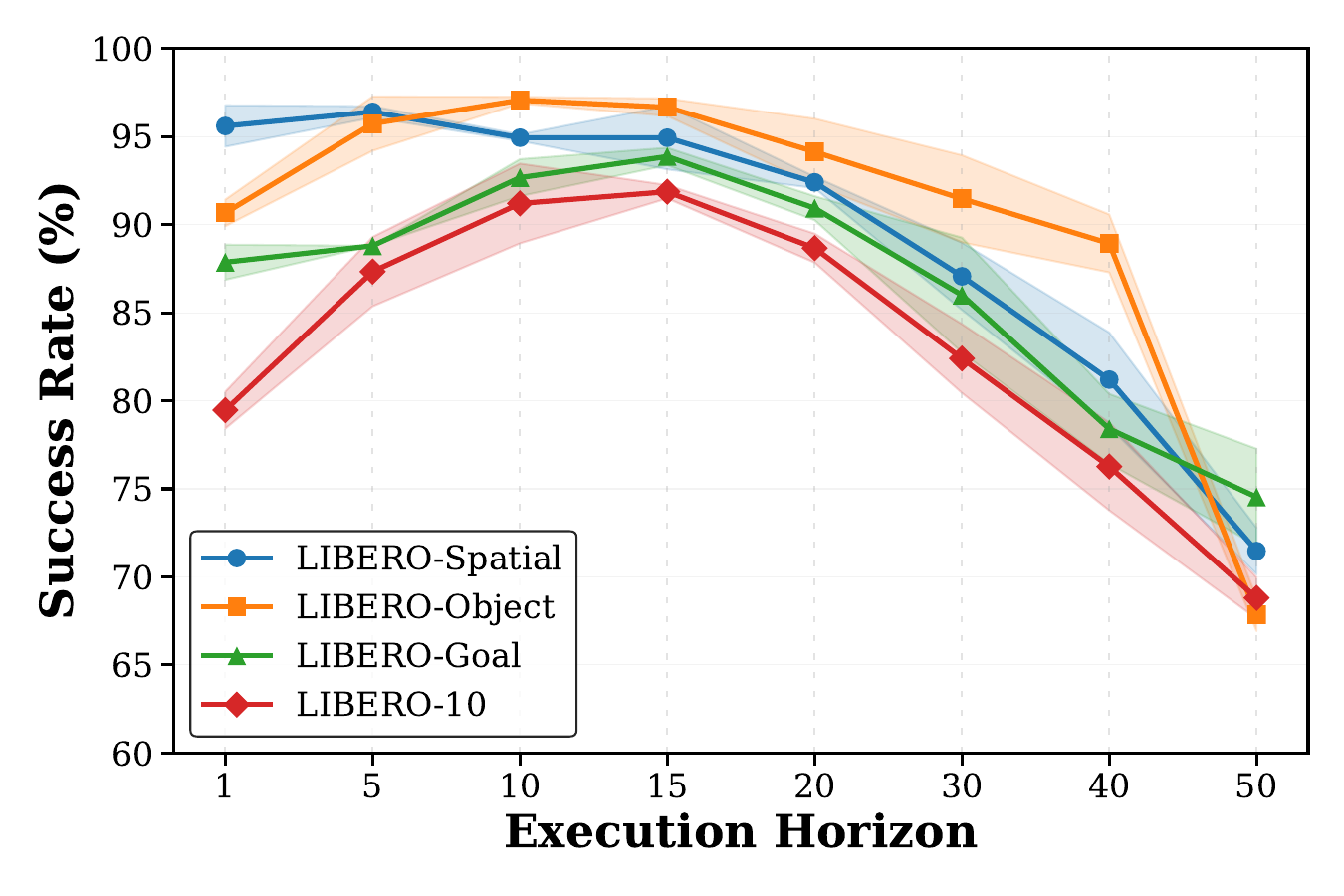}
    \vspace{-20pt}
    \caption{\textbf{Illustration of the average success rates on the LIBERO benchmark using $\pi_{0.5}$.} Varying the execution horizon leads to substantial success rate fluctuations, and the policy performance exhibits a peaked pattern, initially improving and then declining as the execution horizon increases.} %
    \label{fig:pattern}
    \vspace{-18pt}
\end{figure}


To better capture multimodal action distributions and enforce temporal consistency, \textit{action chunking}~\cite{action_chunk,act,dp} has become a standard practice for VLAs trained via imitation learning~\cite{imitation,bc}. Instead of predicting a single action at each step, the policy outputs a sequence of actions---an action chunk. During policy rollout, the robot executes the initial portion (and only rarely the entirety) of the predicted chunk before re-planning, discarding the remaining actions~\cite{dp,pi0}. The total chunk length is termed the \textit{prediction horizon}, and the executed prefix is the \textit{execution horizon}~\cite{rtc}.
This formulation establishes a closed-loop prediction mechanism, sacrificing long-term consistency for reactivity~\cite{bid}.

Empirically, we observed that the policy’s behavior is inherently tied to the length of its executed prefix. As illustrated in Fig.~\ref{fig:pattern}, varying the execution horizon leads to substantial performance fluctuations—ranging from consistent successes to frequent failures. Interestingly, the performance also exhibits a characteristic trend: it initially improves and then declines as the execution horizon increases. These findings underscore the importance of selecting an appropriate execution horizon and motivate a fundamental yet underexplored question in action chunking: \textbf{How should the execution horizon be determined?}


Prior works~\cite{act,dp,pi0,pi05,gr00t} typically set a \textit{fixed} execution horizon, chosen either by human heuristics or through exhaustive evaluation across multiple configurations. Such brute-force tuning quickly becomes time- and compute-intensive as the prediction horizon and task complexity increases.
Furthermore, we argue that a fixed execution horizon is inherently suboptimal.
\citet{bid} showed that short horizons improve reactivity but induce instability due to frequent cross-chunk transitions, whereas long horizons enhance temporal smoothness at the expense of responsiveness.
Since the balance between consistency and reactivity naturally varies across different phases of policy rollout, the optimal execution horizon should likewise adapt over time.
For instance, reaching toward a coffee carafe favors a longer horizon for smooth motion, whereas pouring into a cup requires shorter horizons for heightened responsiveness.
These insights emphasize the need of determining the execution horizon adaptively on a per-chunk basis.

In this paper, we aim to automatically and efficiently determine the execution horizon for each action chunk within flow-based VLAs~\cite{pi05,gr00t}. To explain why the policy performance exhibits the characteristic trend shown in Fig.~\ref{fig:pattern}, we analyze how action generation integrates linguistic and visual information through the attention mechanism~\cite{attention}.
Our analysis reveals two key phenomena: \raisebox{-0.6pt}{\ding[1.1]{182\relax}} 
Intra-chunk actions consistently attend to the same vision–language tokens, indicating that they rely on fixed perceptual contexts. These contexts provide useful guidance for early actions but become increasingly outdated for later ones as the environment changes, revealing the limited adaptability of the predicted chunks.
\raisebox{-0.6pt}{\ding[1.1]{183\relax}} Predicted actions exhibit strong attention to the initial and terminal action tokens, with correspondence strength remaining high before sharply decaying as temporal distance increases. We refer to these boundary tokens as \textit{radial action sinks}, which serve as stable anchors around which intermediate actions are organized.


Building on the observations of the limited adaptability of intra-chunk actions and their reliance on radial action sinks, we formulate execution horizon determination as \textbf{estimating the predictive limit of VLAs}.
We interpret the action self-attention weights as implicit indicators of the model’s prediction confidence and propose \textbf{AutoHorizon}, a dynamic execution horizon estimation strategy for flow-based VLAs trained with action chunking. Our approach leverages the intrinsic structure of attention weights to infer the temporal limit of the model’s reliable forecasting capability. Specifically, we introduce a bidirectional soft-pointer mechanism that locates the first turning points where the attention mass ceases to advance and begins to plateau. These turning points, identified for both initial and terminal radial action sinks, define the estimated execution horizon.

Our contributions are threefold. 
\textbf{(1)} We focus on flow-based VLAs trained with action chunking, and provide both theoretical and empirical analyses of their performance pattern with respect to the execution horizon. We further uncover its underlying causes through two key observations linking attention correspondences to the model’s predictive limit.
\textbf{(2)} Building on these insights, we propose AutoHorizon, a novel attention-guided strategy that dynamically estimates the execution horizon for each action chunk, allowing the policy to adapt to varying perceptual conditions.
\textbf{(3)} Extensive experiments on simulated and real-world robot manipulation tasks demonstrate that our method generalizes across different flow-based policies, incurs negligible computational overhead, 
and outperforms strong baselines that rely on exhaustive fixed-horizon tuning.

%% file: sec/2_related.tex
\section{Related Work}
\label{sec:related}
\subsection{Vision-Language-Action Models}
The remarkable progress of large language models (LLMs)~\cite{gpt4,gemini,qwen3} and vision-language models (VLMs)~\cite{paligemma,eagle25,internvl3} has sparked growing interest in AI systems capable of acting within the physical world. Among these advances, \textit{Vision-Language-Action (VLA) models}~\cite{openvla,pi0,pi05,gr00t,rt2,cotvla,tinyvla,dreamvla,pointvla,3dvla,gr2,uva,vpp} have emerged as a unifying paradigm that integrates perception, reasoning, and control. By jointly training on large-scale datasets of robotic and human demonstrations, VLAs enable end-to-end learning from visual and linguistic inputs to motor actions, demonstrating strong generalization across diverse language-conditioned manipulation tasks.
Within this paradigm, diffusion- and flow-based VLAs~\cite{pi0,pi05,gr00t,smolvla} have received particular attention for their sample-efficient generation process and high-quality action synthesis. Conditioned on the visual and linguistic embeddings extracted from the backbone VLMs~\cite{eagle25,paligemma}, these models iteratively sample from an action distribution to produce low-level motor commands for robotic control, achieving fine-grained temporal consistency and robust performance across complex environments.

\begin{figure*}[t]
    \centering
    \includegraphics[width=1.0\linewidth, trim=0 90 0 60, clip]{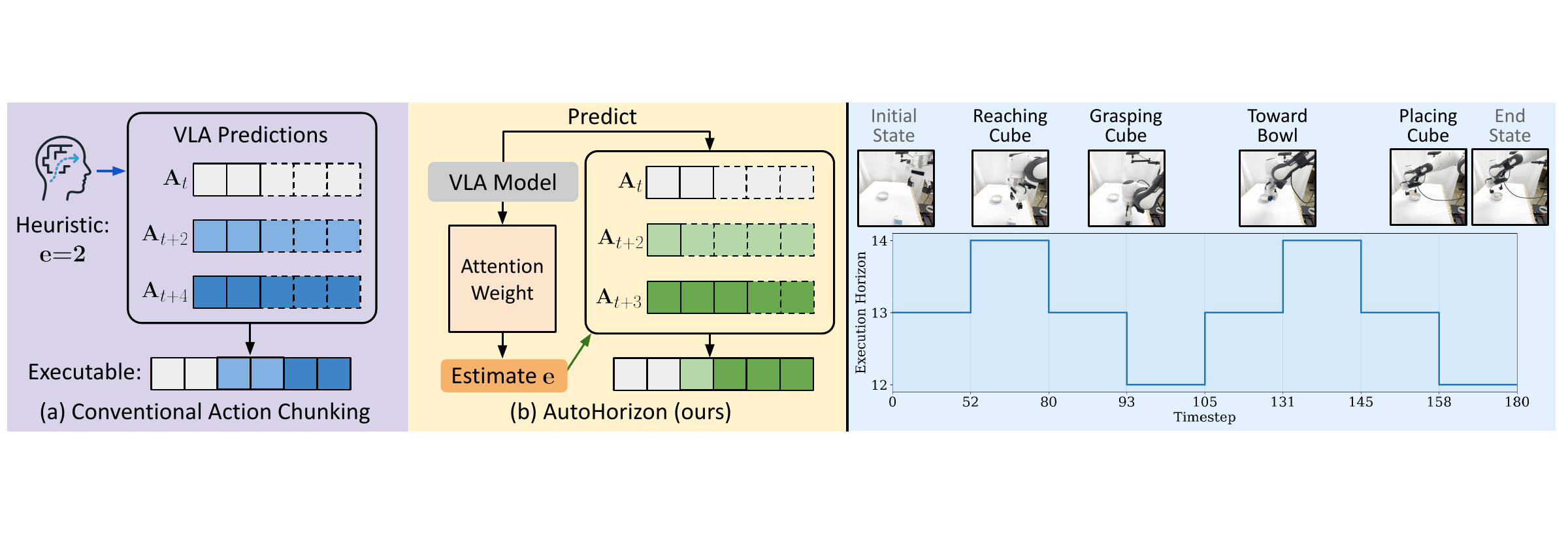}
    \vspace{-18pt}
    \caption{\textbf{Left:} 
    (a) In conventional action chunking, the execution horizon $e$ is heuristically chosen by humans and remains fixed across chunks. 
    (b) In contrast, AutoHorizon (our method) dynamically estimates the execution horizon for each predicted chunk based on the attention weights from the VLA model. 
    \textbf{Right:} 
    Real-world demonstration of showing how the estimated execution horizons evolve during policy rollout. When the environment is stable and reactivity is less critical (\eg, reaching the cube or moving toward the bowl), the estimated horizon increases to promote smooth, stable motion. Conversely, during physical interaction (\eg, grasping or placing the cube), the execution horizon shortens to enhance reactivity and adaptability.
    }
    \label{fig:overview}
    \vspace{-12pt}
\end{figure*}

\subsection{Action Chunking}
Action chunking policies~\cite{act,dp,dp3,pi0,pi05} generate short sequences of consecutive actions for a given state.
ACT~\cite{act} first introduced the idea of action chunking using Transformers~\cite{attention} and demonstrated that modeling temporally consistent action sequences enables the learning of fine-grained behaviors, outperforming policies that predict single actions.
Subsequent works have extended this concept in several directions.
ACT further applied weighted averaging to overlapping actions between chunks to enhance policy smoothness.
BID~\cite{bid} provided a theoretical analysis from a policy learning perspective, showing that action chunking promotes long-term temporal consistency but sacrifices short-term reactivity. 
They also proposed a rejection sampling strategy to select the best-performing chunk to address this trade-off.
RTC~\cite{rtc} explored action chunking under asynchronous execution~\cite{smolvla}, formulating chunk prediction as an image inpainting problem~\cite{pigdm}.

Despite these advances, the \textit{choice} of the execution horizon remains largely unexplored. 
Existing works~\cite{act,dp,pi0} typically set this value using human heuristics, determined by the robot’s control frequency and the desired duration of each motion segment.
This practice has left the influence of the execution horizon on VLA performance largely unexplored, prompting a fundamental question: \textit{are there better ways to determine the execution horizon?}


\subsection{Attention Weights}
Attention weights~\cite{streamingllm,h2o,visual_attnsink,sink_llm} serve as interpretable indicators of information flow between tokens in Transformer-based models~\cite{attention}. By examining self-attention patterns in large language models, StreamingLLM~\cite{streamingllm} observed that LLMs tend to focus disproportionately on initial tokens—a phenomenon termed the \textit{attention sink}. Preserving the corresponding key–value pairs was shown to enable efficient, infinite-length text processing without additional fine-tuning.
Extending this concept to multimodal architectures, \citet{visual_attnsink} found that large vision-language models often assign high attention weights to visually salient but semantically irrelevant tokens. To mitigate this, they proposed redistributing attention away from such tokens to improve cross-modal alignment and visual grounding.

In this work, we analyze the attention weights within flow-based VLAs~\cite{pi05,gr00t}, uncovering key insights into how actions attend to other modalities and therefore propose to interpret action self-attention weights as informative cues for estimating the execution horizon.

%% file: sec/3_method.tex
\section{Methodology}
\label{sec:method}

\begin{figure*}[t]
    \centering
    \includegraphics[width=1.0\linewidth, trim=125 95 140 85, clip]{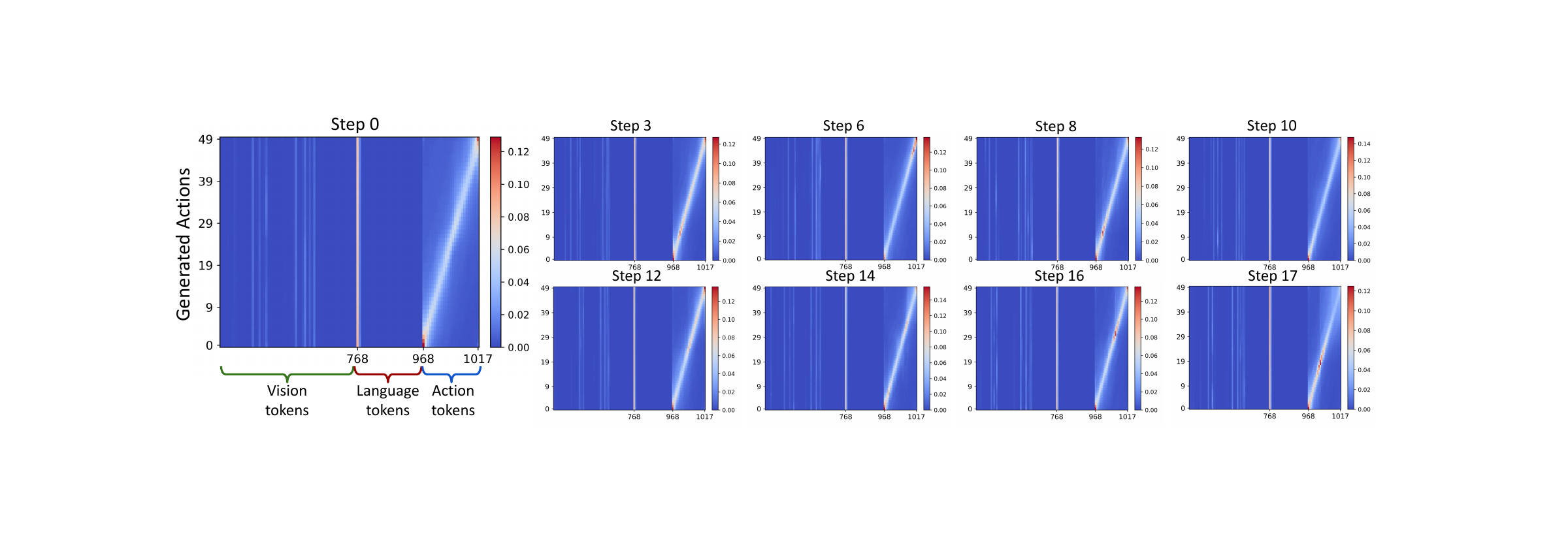}
    \vspace{-18pt}
    \caption{\textbf{Visualization of average attention weights in $\pi_{0.5}$ across different stages of task execution.} Intra-chunk actions consistently attend to the same vision and language tokens across predicted chunks throughout the rollout. This invariance is consistently observed across different sampling steps, task rollouts, and pretrained models. The x-axis is rescaled for clarity of visualization. }
    \label{fig:attn_heatmap}
    \vspace{-12pt}
\end{figure*}

\subsection{Preliminary}
Denote the pretrained diffusion-/flow-based Vision-Language-Action (VLA) model as $\pi(\mathbf{A}_t|\mathbf{o}_t,\mathbf{c})$, where $\mathbf{o}_t$ represents the input visual observations at time step $t$, and $\mathbf{c}$ denotes the corresponding language command. The policy is trained to predict a sequence of $p$ continuous actions,
$\mathbf{A}_t = [\,\mathbf{a}_t,\, \mathbf{a}_{t+1},\, \ldots,\, \mathbf{a}_{t+p-1}\,],$
referred to as an \textit{action chunk}. 
Here, the parameter $p \in \mathbb{N}$ specifies the \textbf{prediction horizon}, \ie, the temporal window over which the model forecasts future actions conditioned on the current perceptual-linguistic context $(\mathbf{o}_t, \mathbf{c})$.
During execution, the agent typically performs the first $e$ actions from the predicted chunk before re-sampling new input observations and generating the next action chunk, where $e \in \mathbb{N}$ defines the \textbf{execution horizon}.

Attention weights are widely employed to quantify the correspondence between different tokens. 
Let $T_{v}$, $T_{l}$, and $T_{a}$ denote the number of encoded vision, language, and action tokens within the VLA respectively. 
Within a standard transformer-based attention mechanism, the attention weight matrix $\mathbf{S}$ is defined as the post-softmax similarity between the query and key embeddings:
\begin{equation}
\label{eq:attn_score}
    \mathbf{S} = \text{softmax}\!\left(\frac{\mathbf{Q}\mathbf{K}^{\top}}{\sqrt{d}}\right),
\end{equation}
where $\mathbf{Q} \in \mathbb{R}^{T_a \times d}$ represents the queries derived from action tokens, 
$\mathbf{K} \in \mathbb{R}^{(T_{v}+T_{l}+T_{a}) \times d}$ denotes the concatenated keys from vision, language, and action tokens, 
and $d$ is the shared token feature dimension. Each entry $\mathbf{S}_{ij}$ captures the relative attention weight between the $i$-th action query and the $j$-th key token, 
thus encoding how strongly the policy attends to specific visual regions, linguistic cues, or consecutive actions when generating its action sequence.

In the following, we first show that flow-matching policy performance exhibits a peaked trend with respect to the execution horizon, underscoring the need and feasibility to identify an optimal value.
To explain this phenomenon, we analyze the attention weights and reveal two key phenomena that shape predicted chunk behavior.
Building on these insights, we introduce an efficient strategy for execution horizon estimation.
Unless otherwise specified, all analyses are conducted using the state-of-the-art $\pi_{0.5}$~\cite{pi05} model.

%


\subsection{Existence of Optimal Execution Horizon}
\label{sec:opt_e}
Consider an action chunking policy that employs a fixed execution horizon of length $e$ throughout its rollout. 
Assume the policy executes a total of $L$ low-level actions before task termination. 
Let $\delta^{c}$ denote the loss in final task reward incurred at each chunk transition, assumed to be independent of $e$. 
Let $\delta^{d}_{j}(e)$ represent the total divergence loss between the $j$-th executed action chunk and its corresponding expert trajectory segment. With $m=\left\lceil L/e\right\rceil$ executed chunks in total, the expected error accumulated over the rollout can be expressed as: 
\begin{equation}
\label{eq:total_loss}
\mathcal{L}(e) = 
\sum_{i=0}^{m -1} \delta^{c}
\;+\;
\sum_{j=0}^{m} \delta^{d}_{j}(e).
\end{equation}

\begin{prop}[Unique Error Minimizer]
Suppose $\delta^{d}_{j}(e)$ is a monotonically increasing function with respect to $e$ and can be modeled as $\delta^{d}_{j}(e) = k e \log e$, where $k > 0$ is a scaling factor. Denote $p$ as the prediction horizon. 
Assuming $L$ is divisible by $e$, there exists a unique minimizer of $\mathcal{L}(e)$:
\begin{equation}
\label{eq:min_e}
e^{*} \;=\; 
\mathrm{clamp}\!\left(
\Big\lceil \tfrac{\delta^{c}}{k} \Big\rceil 
\;\text{ or }\;
\Big\lceil \tfrac{\delta^{c}}{k} \Big\rceil - 1, 1,\, p
\right),
\end{equation}
such that $\mathcal{L}(e)$ is strictly decreasing on $(0, e^{*})$ and strictly increasing on $(e^{*}, \infty)$.
\end{prop}
Proof is provided in Sec.~\ref{sec:proof}. Eq.~\eqref{eq:min_e} indicates that when the policy is trained on a diverse set of the underlying action distributions and the chunk transition loss $\delta^{c}$ is large, a longer execution horizon is preferred to promote temporally consistent action sequences.
Conversely, when the policy struggles to accurately model the implicit environment dynamics and the intra-chunk divergence loss $\delta^{d}(e)$ dominates, a shorter execution horizon becomes more favorable, enhancing reactivity to environmental changes.
Our analysis can be viewed as an extension of \citet{bid}. By modeling the total policy rollout error, we demonstrate that an optimal execution horizon exists and that performance monotonically decreases as the horizon deviates from this optimum, thereby establishing an \textit{explicit} trade-off between long-term consistency and short-term reactivity. Note that Eq.~\ref{eq:min_e} serves only as a theoretical proof of existence and does not directly guide method design.



Empirically, we also observe the characteristic relationship between policy performance and execution horizon. 
As shown in Fig.~\ref{fig:pattern}, varying the execution horizon leads to substantial performance deviations, underscoring the importance of selecting an appropriate value of $e$. 
Across all evaluated tasks, the behavior of the policy performance aligns with our theoretical findings: the success rate initially increases and then declines as $e$ grows, peaking at an intermediate value. This indicates that optimal performance arises from balancing consistency with reactivity.
Fig.~\ref{fig:supp_h10} further illustrates the case with a smaller prediction horizon ($p = 10$). In this constrained setting, the best performance typically occurs at the horizon boundary, i.e., $e^{*} = p$. This behavior arises because the policy is well-trained under a small $p$, allowing it to accurately capture the implicit environment dynamics. As a result, the predicted trajectories closely match the expert demonstrations, leading to $k \rightarrow 0$. 
However, once the execution horizon exceeds the prediction horizon ($e > p$), a pronounced performance drop emerges. We attribute this to a train–test mismatch, since the model has not been trained on trajectories longer than $p$.

\subsection{VLA Knows Its Limits}
To uncover the underlying causes of the observed performance pattern, we analyze the model’s attention weights to examine how the VLA allocates focus across vision, language, and action tokens during action generation.
By interpreting these attention distributions, we identify two key phenomena that jointly explain the empirical patterns reported in Sec.~\ref{sec:opt_e} and shed light on how attention dynamics reflect the VLA’s intrinsic predictive limits.

\noindent\textbf{\raisebox{-0.6pt}{\ding[1.1]{182\relax}} Intra-chunk actions attend invariantly to vision–language tokens.} 
We visualize the cross-attention maps of the last sampling step from the $\pi_{0.5}$ model ($p = 50$) in Fig.~\ref{fig:attn_heatmap}.
Each row in the heatmap represents the attention distribution between action queries and the concatenated sequence of vision, language, and action keys.
The first 768 tokens correspond to visual input, followed by 200 language tokens, and the remaining correspond to action tokens.
Column widths are rescaled for improved clarity.

From the figure, we observe a striking invariance: actions within the same chunk consistently attend to the same vision–language tokens with nearly identical importance.
This suggests that, although the model predicts a temporally extended sequence of future actions, later actions fail to adaptively adjust to environmental changes.
Instead, they repeatedly rely on static visual–linguistic features that are informative for early actions but increasingly outdated or misleading for later ones.
Consequently, executing these later actions may become redundant or even detrimental, contributing little new contextual information and corrective flexibility.
This property manifests as overconfident policy rollouts with reduced reactivity and adaptability.

We also observe an unusually high concentration of attention on the \textit{first language token}, a phenomenon that persists across all transformer blocks and sampling steps.
This pattern resembles the attention sink effect reported in large language models~\cite{streamingllm}.
However, unlike in LLMs, where sink tokens often encode structural or positional information, the language attention sink in VLAs appears largely redundant and carries minimal semantic value.
Empirically, masking all language tokens (by setting their attention weights to zero) may results in only a marginal drop in task success rate (Sec.~\ref{sec:lang_effect}).
We infer that, due to the strong vision–language pretraining of the backbone model, most linguistic semantics are already embedded within the visual representations during action generation, making explicit linguistic attention largely superfluous.

\begin{figure}[t]
    \centering
    \includegraphics[width=1.0\linewidth, trim=338 130 345 90, clip]{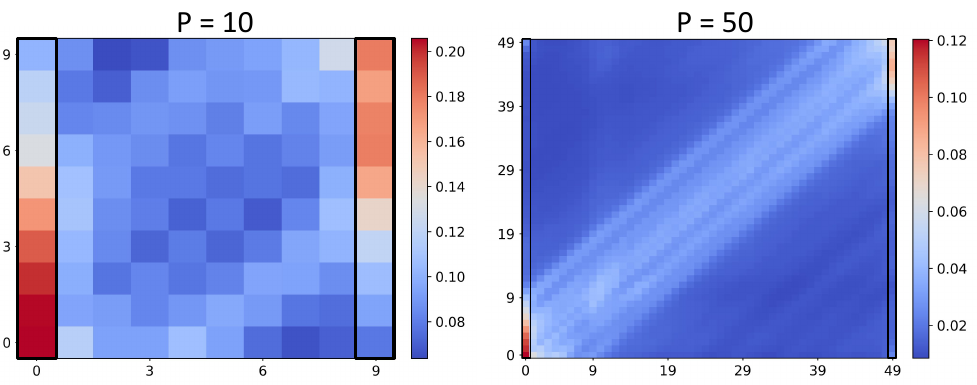}
    \vspace{-22pt}
    \caption{\textbf{Visualization of normalized action self-attention weights.} 
    Across different prediction horizons, the predicted actions exhibit strong attention to the initial and terminal action tokens, with correspondence strength remaining high before sharply decaying as temporal distance increases. 
    These boundary tokens (encircled in black) are referred to as \textit{radial action sinks}.}
    \label{fig:self_attn_heatmap}
    \vspace{-18pt}
\end{figure}

\noindent\textbf{\raisebox{-0.6pt}{\ding[1.1]{183\relax}} Radial action sinks emerge at the sequence boundaries.} 
In Fig.~\ref{fig:self_attn_heatmap}, we visualize the normalized self-attention maps among action tokens under varying prediction horizons.
A consistent pattern emerges: strong attention is concentrated on both the initial and terminal action tokens.
The correspondence strength remains high for a short span, then rapidly decays over a few actions before stabilizing into a low-attention plateau.
We refer to these highly attended boundary tokens as \textbf{radial action sinks}.

Why do flow-based VLAs consistently concentrate attention at the beginning and end of each predicted chunk?
We infer two underlying reasons.
First, the initial action tends to exhibit the lowest cumulative error and thus serves as a stable anchor for subsequent actions.
Later actions follow this anchor’s guidance while gradually attending more to adjacent tokens, resulting in smooth and temporally consistent trajectories.
Second, because the policies are trained on expert demonstrations with randomly sampled starting timestamps, both the initial and terminal actions play a role in preserving inter-chunk continuity, reflecting the model’s implicit objective to ensure smooth transitions across chunk boundaries.
Together, these radial action sinks establish an attention structure in which the initial and terminal tokens define latent centers around which intermediate actions are organized, revealing an intrinsic bias in the model’s temporal reasoning process.

\noindent\textbf{Action self-attention as an indicator of predictive limit.}
Building on these two observations, we conclude that intra-chunk actions exhibit limited adaptability and attend strongly to radial action sinks.
Therefore, we \textit{pose to interpret the action self-attention weights as implicit indicators of the model’s predictive limit}.
Specifically, when the self-attention weights associated with the radial action sinks remain high, the model is confident that its predicted actions are still aligned with the guiding anchors and thus valid under the current observation.
In contrast, as these attention weights decline, the model increasingly conditions on its own previously generated actions rather than grounded sensory inputs.
This shift toward self-referential dependence amplifies compounding errors and ultimately degrades reactivity and performance during extended rollouts.


\begin{table*}[t]
\centering
\small
\caption{\textbf{Performance comparison of $\pi_{0.5}$ on LIBERO benchmark under different prediction horizons.}  Best results are in \textbf{bold}.}
\vspace{-8pt}
\adjustbox{width=1.0\textwidth}{
\begin{tabular}{cccccc|cccc}
\toprule
\multicolumn{2}{c|}{Setting}                                           & \multicolumn{4}{c|}{$p=10$}                                & \multicolumn{4}{c}{$p=50$}                                 \\ 
\midrule
\multicolumn{2}{c|}{Task Suite}                                                                                                   & LIB-Spatial & LIB-Object & LIB-Goal & LIB-10 & LIB-Spatial & LIB-Object & LIB-Goal & LIB-10 \\ \midrule
\multicolumn{1}{c|}{\multirow{5}{*}{\begin{tabular}[c]{@{}c@{}}Static \\ Oracle\end{tabular}}} & \multicolumn{1}{c|}{$e = 0.2p$} & 98.5 $\pm$ 0.5 & 94.1 $\pm$ 1.0 & 90.4 $\pm$ 0.0 & 76.0 $\pm$ 1.2 & 94.9 $\pm$ 0.2 & 97.1 $\pm$ 0.2 & 92.7 $\pm$ 1.0 & 91.2 $\pm$ 2.3 \\
\multicolumn{1}{c|}{} & \multicolumn{1}{c|}{$e=0.4p$} & 98.9 $\pm$ 0.4  & 98.1 $\pm$ 0.5 & 94.8 $\pm$ 0.9 & 82.8 $\pm$ 1.5  & 92.4 $\pm$ 0.3 & 94.1 $\pm$ 1.9 & 90.9 $\pm$ 0.7 & 88.7 $\pm$ 0.8 \\
\multicolumn{1}{c|}{}  & \multicolumn{1}{c|}{$e = 0.6p$} & 98.8 $\pm$ 0.3 & 98.7 $\pm$ 0.7 & 95.2 $\pm$ 0.7 & 85.9 $\pm$ 0.8  & 87.1 $\pm$ 1.9 & 91.5 $\pm$ 2.5 & 86.0 $\pm$ 3.3 & 82.4 $\pm$ 2.0 \\
\multicolumn{1}{c|}{} & \multicolumn{1}{c|}{$e = 0.8p$} & 98.9 $\pm$ 0.4 & 99.1 $\pm$ 0.5 & 95.1 $\pm$ 1.1 & 88.5 $\pm$ 1.5 & 81.2 $\pm$ 2.7 & 88.9 $\pm$ 1.6 & 78.4 $\pm$ 2.0 & 76.3 $\pm$ 2.5 \\
\multicolumn{1}{c|}{} & \multicolumn{1}{c|}{$e = 1.0p$} & 99.1 $\pm$ 0.5  & 98.8 $\pm$ 0.9 &  97.2 $\pm$ 1.0  & 90.4 $\pm$ 0.6 & 71.5 $\pm$ 1.3 & 67.9 $\pm$ 0.9 & 74.5 $\pm$ 2.7 & 68.6 $\pm$ 1.2 \\ \midrule
\multicolumn{2}{c|}{Static Oracle+}   & 99.1 $\pm$ 0.5  & 99.1 $\pm$ 0.5 & 97.2 $\pm$ 1.0 & 90.4 $\pm$ 0.6 & 96.4 $\pm$ 0.3  & 97.6 $\pm$ 0.6 & 93.9 $\pm$ 0.5 & 91.9 $\pm$ 0.4 \\ \midrule
\multicolumn{2}{c|}{Random} & 98.4 $\pm$ 1.2 & 98.4 $\pm$ 0.7 & 96.3 $\pm$ 0.7 & 86.3 $\pm$ 1.5 & 82.8 $\pm$ 1.5 & 90.3 $\pm$ 1.6  & 85.3 $\pm$ 2.3 & 83.3 $\pm$ 0.7 \\ \midrule
\multicolumn{2}{c|}{\textbf{AutoHorizon}} &  \textbf{99.1 $\pm$ 0.2}  & \textbf{99.2 $\pm$ 0.3} & \textbf{97.5 $\pm$ 0.2} & \textbf{91.6 $\pm$ 0.7} & \textbf{96.5 $\pm$ 0.9} & \textbf{98.0 $\pm$ 0.6} & \textbf{94.4 $\pm$ 1.0} & \textbf{92.1 $\pm$ 1.0} \\ \bottomrule
\end{tabular}
}
\label{tab:libero_pi05}
\vspace{-12pt}
\end{table*}

\subsection{AutoHorizon}
Motivated by the above analysis, we propose leveraging attention weights as a proxy to estimate the execution horizon for \textit{each} action chunk.
Adopting a per-chunk horizon is intuitive: as task difficulty and environmental dynamics fluctuate throughout the policy rollout, the optimal execution horizon should adapt correspondingly to maintain an effective balance between temporal consistency and reactivity.

To this end, we introduce \textbf{AutoHorizon}—a data-adaptive approach that estimates execution horizons directly from the model’s intrinsic attention dynamics.
Given the attention weights obtained at each sampling step $t$, we first extract the action self-attention maps and average them across all transformer blocks and attention heads, followed by row-wise normalization to ensure that each query–key distribution sums to one.
The resulting normalized attention matrix is denoted as
$
\mathbf{S}_t \in \mathbb{R}^{p\times p},
$
where $p$ is the prediction horizon, and notations are reused from Eq.~\eqref{eq:attn_score} for simplicity.
Intuitively, $\mathbf{S}_t[i,j]$ quantifies how strongly the $i$-th query action attends to the $j$-th key action, revealing how far the model effectively “looks ahead.”
Our objective is to identify the first turning point in the attention trajectory—where the attention mass stops advancing and begins to plateau—which marks the natural boundary of the VLA’s predictive limit, ensuring that all executed actions remain reliable while maximizing temporal consistency.

To estimate this turning point, we first retain only the rows of $\mathbf{S}_t$ that exhibit low entropy, defined as
\begin{equation}
\label{eq:param_q}
R_t = \{\, i \mid H_t[i] \le Q_q(H_t) \,\},
\end{equation}
where 
\begin{equation}
H_t[i] = -\frac{1}{\log p}\sum_{j}\mathbf{S}_t[i,j]\log\mathbf{S}_t[i,j],    
\end{equation}
and $Q_q$ denotes the $q$-quantile of the entropy distribution across rows. 
This filtering step removes actions with uniformly diffused attention, preserving those with sharper, more confident patterns that provide reliable structural cues.

Next, we employ a bidirectional soft-pointer mechanism to locate the plateau.
Two pointers, $q_s = 0$ and $q_e = p - 1$, are initialized at the start and end of the chunk, respectively.
For the forward pointer $q_s$, the expected predictive horizon for each row $i$ is computed as
\begin{equation}
\label{eq:expected_n}
\mu_t[i] = \max\!\left( \sum_{j=0}^{p-1} j\,\mathbf{S}_t[i,j],\, \max_{k \le i}\mu_t[k] \right),
\end{equation}
where the non-decreasing constraint enforces monotonic progression and prevents backward jumps.
We then compute the incremental change $\Delta\mu_t[i] = \mu_t[i] - \mu_t[i-1]$, which tracks the evolution of the attention trajectory.
A large $\Delta\mu_t[i]$ indicates a sudden shift in attention focus, signaling the onset of a plateau.
The set of actions preceding this plateau is defined as
\begin{equation}
\label{eq:plateau}
P_t = \{\, i \mid \Delta\mu_t[i] < \tau \,\},
\end{equation}
where $\tau$ is a fixed threshold.
At sampling step $t$, the forward execution horizon is then determined as
\begin{equation}
\label{eq:N_f}
N_f = \left\lfloor \mu_t\!\left[\min(R_t \cap P_t)\right] \right\rfloor + 1.
\end{equation}
The same procedure is applied to the reversed attention matrix $\tilde{\mathbf{S}}_t$ to obtain the backward horizon $N_b$. 
If the combined coverage satisfies $N_f + N_b \ge p$, a full-range coverage is adopted ($N = p$); otherwise, only the forward prefix length is used as the effective execution horizon ($N = N_f$). 
Empirically, we find that the former case typically occurs when the prediction horizon $p$ is small, whereas the latter dominates for larger $p$.
A concrete example illustrating the full procedure is shown in Fig.~\ref{fig:method_example}.


%% file: sec/4_experiment.tex
\section{Experiments}
\subsection{Experimental Settings}
\noindent\textbf{Models.}
We evaluate our method on two representative flow-based Vision–Language–Action models: $\pi_{0.5}$~\cite{pi05} and GR00T~N1.5~\cite{gr00t}. 
For $\pi_{0.5}$, we conduct experiments with two variants using prediction horizons of $p = 10$ and $p = 50$ to examine horizon-dependent behavior.
For GR00T~N1.5, we adopt the publicly released pretrained checkpoints with the default prediction horizon of $p = 16$.
AutoHorizon operates on the first or third sampling step and typically uses fixed hyperparameters of $q = 0.9$ and $\tau = 0.3$, requiring no additional parameter tuning.

\noindent\textbf{Baselines.}
We compare against the following baselines:
\begin{itemize} 
    \item \textbf{Static Oracle.}  
    This corresponds to the conventional setting in action chunking policies, where a fixed execution horizon is maintained throughout rollout. Execution horizons are uniformly sampled for full range coverage.
    \item \textbf{Static Oracle+.}  
    An enhanced version of Static Oracle that performs brute-force search over the prediction horizon, thereby achieving optimal performance under fixed horizon settings. It serves as a strong yet costly baseline, as it requires $p$ rollouts per task. 
    \item \textbf{Random:} 
    To assess the effect of adaptive horizon selection, we include a stochastic baseline where the execution horizon is randomly sampled at each rollout step, following $e \sim \mathcal{U}(1, p)$.  
    This baseline isolates the contribution of structured, attention-guided adaptation from performance variations arising purely from random horizon changes.
\end{itemize}
For all experiments, we report both the mean and standard deviation to ensure fair comparison and robust evaluation. We also compare against two adaptive re-planning baselines, Action Trigger and Uncertainty Proxy, in Sec.~\ref{sec:additional_res}. Although these baselines are not central to the focus of this paper, we include them to further demonstrate the effectiveness of AutoHorizon.

\begin{table}[t]
\centering
\small
\caption{\textbf{Performance comparison using GR00T~N1.5 on the LIBERO benchmark.}  Best results are highlighted in \textbf{bold}.}
\vspace{-8pt}
\adjustbox{width=1.0\linewidth}{
\begin{tabular}{cc|cccc}
\toprule
\multicolumn{2}{c|}{Task Suite} & LIB-Spatial & LIB-Object & LIB-Goal & LIB-10 \\ \midrule
\multicolumn{1}{c|}{\multirow{5}{*}{\begin{tabular}[c]{@{}c@{}}Static \\ Oracle\end{tabular}}} & $e=1$ & 92.7 $\pm$ 0.9  &   94.7 $\pm$ 3.4 & 82.7 $\pm$ 0.9 & 74.7 $\pm$ 3.4 \\
\multicolumn{1}{c|}{}  & $e=2$ & 96.0 $\pm$ 1.6  & 95.3 $\pm$ 3.8 & 82.0 $\pm$ 1.6 & 83.3 $\pm$ 0.9 \\
\multicolumn{1}{c|}{} & $e=4$ & 94.7 $\pm$ 3.4 & 95.3 $\pm$ 2.5 & 90.7 $\pm$ 0.9 & 88.7 $\pm$ 1.9 \\
\multicolumn{1}{c|}{} & $e=8$ & 95.3 $\pm$ 0.9 & 97.3 $\pm$ 0.9 & 94.7 $\pm$ 3.4 & 86.0 $\pm$ 5.9 \\
\multicolumn{1}{c|}{}  & $e=12$ & 91.3 $\pm$ 2.5 & 96.0 $\pm$ 0.0 & 90.0 $\pm$ 0.0  & 86.0 $\pm$ 4.3 \\
\multicolumn{1}{c|}{}  & $e=16$ & 94.0 $\pm$ 3.3 & 94.7 $\pm$ 0.9 & 90.7 $\pm$ 0.9 & 90.0 $\pm$ 1.6 \\ \midrule
\multicolumn{2}{c|}{Static Oracle+}   & 96.0 $\pm$ 1.6 & 97.3 $\pm$ 0.9 & 94.7 $\pm$ 3.4 & 90.0 $\pm$ 1.6 \\ \midrule
\multicolumn{2}{c|}{Random} & 93.3 $\pm$ 0.9 & 96.0 $\pm$ 2.8 & 92.0 $\pm$ 1.6 & 88.7 $\pm$ 1.9 \\ \midrule
\multicolumn{2}{c|}{\textbf{AutoHorizon}} & \textbf{96.7 $\pm$ 0.9} & \textbf{98.7 $\pm$ 1.8} & \textbf{96.0 $\pm$ 1.6} & \textbf{92.7 $\pm$ 2.5} \\ \bottomrule
\end{tabular}
}
\label{tab:libero_gr00t}
\vspace{-12pt}
\end{table}

\subsection{Simulation Results}
We evaluate AutoHorizon in simulated robotic manipulation environments that require both short- and long-horizon decision making.
Our experiments leverage two benchmark datasets: the LIBERO dataset~\cite{libero}, which offers a diverse suite of single-arm manipulation tasks, and the RoboTwin dataset~\cite{robotwin,robotwin2.0}, which focuses on bimanual coordination tasks.
For LIBERO, we include four subsets—\textit{LIBERO-Spatial}, \textit{LIBERO-Goal}, \textit{LIBERO-Object}, and \textit{LIBERO-10}—and perform $25$ independent rollouts per task. 
For RoboTwin, we evaluate on seven manipulation tasks:
\textit{adjust bottle position},
\textit{pick dual bottles},
\textit{place container onto plate},
\textit{stack two bowls}, 
\textit{place empty cup on coaster}, 
\textit{open laptop}, and
\textit{press stapler}.
Each task is executed for 100 trials.
All experiments are repeated three times to account for stochasticity in action sampling and environment initialization, ensuring statistical robustness.

\noindent\textbf{LIBERO Benchmark.}
Tab.~\ref{tab:libero_pi05} reports the results of $\pi_{0.5}$ on the LIBERO benchmark.
Under a small prediction horizon ($p = 10$), the optimal execution horizon for the \textit{Static Oracle} baseline typically appears at the upper bound of valid values.
The \textit{Random} baseline also performs relatively well, suggesting that models trained with short prediction horizons tend to overfit and accurately capture short trajectory segments.
When the prediction horizon increases to $p = 50$, we observe a significantly different behavior. 
The performance of \textit{Static Oracle} first rises and then declines as the execution horizon extends, while the \textit{Random} baseline suffers a pronounced drop in performance.
In many cases, suboptimal horizon choices even cause \textit{Static Oracle} to underperform the \textit{Random} baseline, underscoring the importance of selecting an appropriate execution horizon.
The enhanced \textit{Static Oracle+} consistently achieves strong results, and the specific horizon values used for this baseline are listed in Sec.~\ref{sec:static_plus}. 
Across both horizon configurations, AutoHorizon consistently outperforms all baselines. 
We attribute this improvement to its ability to dynamically adapt execution horizons during rollout, effectively balancing long-term consistency with short-term reactivity.

\begin{figure}[t]
    \centering
    \includegraphics[width=1.0\linewidth, trim=0 5 0 6, clip]{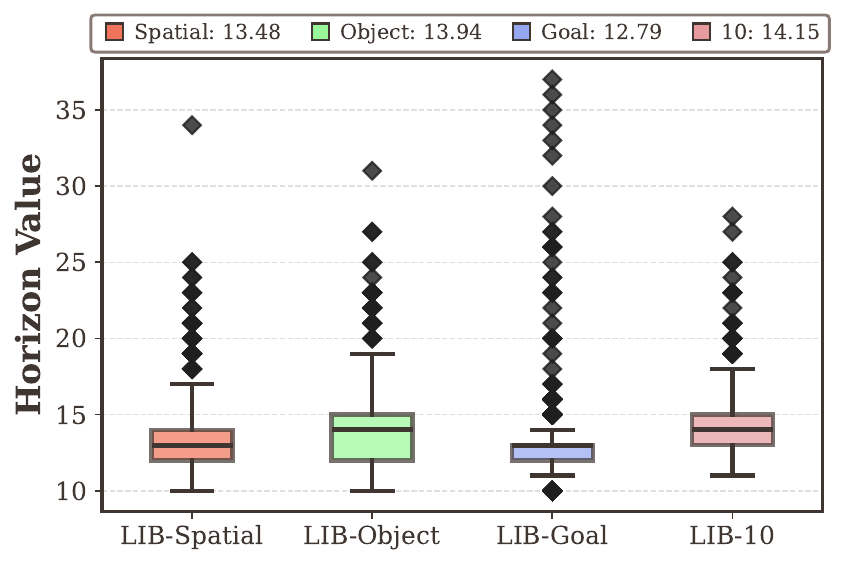}
    \vspace{-18pt}
    \caption{\textbf{Estimated execution horizon distributions by AutoHorizon.} The legend displays the mean values of the distributions.}
    \label{fig:boxplot}
    \vspace{-12pt}
\end{figure}

We further evaluate AutoHorizon on the LIBERO benchmark using GR00T~N1.5~\cite{gr00t}, which differs from $\pi_{0.5}$ in both architectural design and training pipeline.
As shown in Tab.~\ref{tab:libero_gr00t}, although the precise performance trends vary, \textit{Static Oracle} again exhibits a characteristic peak followed by degradation as the execution horizon increases.
Our method consistently achieves superior results, demonstrating robustness and generalization across different architectures and training regimes. In Sec.~\ref{sec:gr00t}, we also visualize the attention weight distributions for GR00T~N1.5, which yield conclusions consistent with our earlier analysis.

\begin{table*}[t]
\centering
\small
\caption{\textbf{Performance comparison using $\pi_{0.5}$ on the RoboTwin tasks.}  Best results are highlighted in \textbf{bold}.}
\vspace{-8pt}
\adjustbox{width=1.00\linewidth}{
\begin{tabular}{cc|ccccccc}
\toprule
\multicolumn{2}{c|}{Task Suite} & Adjust Bottle & Pick Bottles & Place Container & Stack Bowls & Place Cup & Open Laptop & Press Stapler \\ \midrule
\multicolumn{1}{c|}{\multirow{5}{*}{\begin{tabular}[c]{@{}c@{}}Static \\ Oracle\end{tabular}}} & $e=0.2p$ & 79.0 $\pm$ 1.4 & 40.7 $\pm$ 0.5 & 84.0 $\pm$ 2.8 & 90.7 $\pm$ 1.2 & 83.7 $\pm$ 0.9 & 68.3 $\pm$ 1.9 & 44.0 $\pm$ 0.0 \\
\multicolumn{1}{c|}{}  & $e=0.4p$ & 89.0 $\pm$ 0.0 & 67.0 $\pm$ 0.0 & 83.0 $\pm$ 0.0 & 87.3 $\pm$ 1.7 & 77.7 $\pm$ 2.5 & 78.3 $\pm$ 0.5 & 48.0 $\pm$ 0.0 \\
\multicolumn{1}{c|}{} & $e=0.6p$ & 89.3 $\pm$ 0.9 & 65.0 $\pm$ 0.0 & 82.0 $\pm$ 0.0 & 86.7 $\pm$ 2.5 & 68.3 $\pm$ 0.5 & 71.7 $\pm$ 2.1 & 67.0 $\pm$ 0.0 \\
\multicolumn{1}{c|}{} & $e=0.8p$ & 76.7 $\pm$ 1.2 & 58.0 $\pm$ 1.4 & 81.0 $\pm$ 0.0 & 90.0 $\pm$ 2.8 & 66.0 $\pm$ 1.6 & 78.7 $\pm$ 0.5 & 70.0 $\pm$ 0.0 \\
\multicolumn{1}{c|}{}  & $e=1.0p$ & 58.7 $\pm$ 1.2 & 30.0 $\pm$ 0.0 & 76.7 $\pm$ 0.5 & 87.7 $\pm$ 1.2 & 56.3 $\pm$ 3.8 & 84.0 $\pm$ 0.0 & 71.0 $\pm$ 0.0 \\ \midrule
\multicolumn{2}{c|}{Static Oracle+}  & 98.7 $\pm$ 0.5 & 67.0 $\pm$ 0.0 & 91.0 $\pm$ 0.8 & 90.7 $\pm$ 1.2 & 83.7 $\pm$ 0.9 & 84.0 $\pm$ 0.0 & 72.0 $\pm$ 0.0 \\ \midrule
\multicolumn{2}{c|}{Random} & 85.3 $\pm$ 0.5 & 60.0 $\pm$ 0.0 & 86.0 $\pm$ 0.0 & 88.7 $\pm$ 3.4 & 70.7 $\pm$ 2.5 & 82.0 $\pm$ 0.0 & 69.0 $\pm$ 0.0 \\ \midrule
\multicolumn{2}{c|}{\textbf{AutoHorizon}} & \textbf{100.0 $\pm$ 0.0} & \textbf{68.0 $\pm$ 0.0} & \textbf{91.0 $\pm$ 0.8} & \textbf{92.0 $\pm$ 0.8} & \textbf{85.3 $\pm$ 2.1} & \textbf{84.7 $\pm$ 0.9} & \textbf{75.0 $\pm$ 0.0} \\ \bottomrule
\end{tabular}
}
\label{tab:robotwin_pi05}
\vspace{-12pt}
\end{table*}

\noindent\textbf{RoboTwin Benchmark.} 
Tab.~\ref{tab:robotwin_pi05} presents the results on the RoboTwin benchmark across tasks with varying difficulty. 
For both tasks that are highly sensitive (\textit{pick bottles}) and those that are relatively insensitive (\textit{stack bowls}) to the choice of execution horizon, our method achieves comparable or superior performance to the baselines, demonstrating its adaptability across diverse task dynamics.

\noindent\textbf{Ablation.}
Fig.~\ref{fig:boxplot} visualizes the value distribution of execution horizons estimated by AutoHorizon across the four LIBERO task suites.
AutoHorizon yields a broad range of horizon lengths during rollout, demonstrating adaptability to diverse input conditions and capturing the frequent shifts in VLA's prediction dynamics.
Most estimated horizons fall within moderately low values—favoring reactivity—while occasional larger horizons facilitate faster task completion when long-term consistency is beneficial.
We further compare with a variation of \textit{Static Oracle} using fixed horizons closest to AutoHorizon's mean estimated values (\eg, $e = 14$ and $e = 15$ for LIBERO-10) in Sec.~\ref{sec:additional_res}, and find that AutoHorizon consistently achieves higher success rates.
These results confirm the effectiveness of dynamic horizon adjustment over fixed-horizon strategies.

We also examine the effect of hyperparameters in Sec.~\ref{sec:hyper_abl}.
The results show that AutoHorizon's performance remains stable across different combinations of parameter settings.
Compared with the strong \textit{Static Oracle+} baseline, it always achieves comparable or even superior results, demonstrating robustness to hyperparameter choices.

\subsection{Real-World Results}
\noindent\textbf{Setup.}
We further evaluate AutoHorizon in real-world robotic manipulation scenarios.
Experiments are conducted on a Franka Research~3 robot (7-DoF arm)~\citep{franka} following the DROID experimental setup~\citep{droid}.
The backbone VLA is $\pi_{0.5}$, configured with a prediction horizon of $p = 50$.

We assess all methods on three single-arm pick-and-place tasks of increasing difficulty:
\textit{put cucumber on plate},
\textit{put Rubik’s cube on plate}, and
\textit{put Rubik’s cube into bowl}.
A total of 150 trajectories are collected for model fine-tuning.
For evaluation, we adopt a stage-based solve rate that measures progress across four phases:
(1) reaching the object,
(2) grasping and lifting it,
(3) moving it toward the target location, and
(4) successfully placing it in the container.
Each task is evaluated over ten trials per setting, with each trial capped at 300 control steps, amounting to approximately three hours of total robot execution time.
Object positions and orientations are randomized across trials to ensure robustness and generalization.

\begin{table}[t]
\centering
\small
\caption{\textbf{Performance comparison on real-world tasks.}  Best results are highlighted in \textbf{bold}.}
\vspace{-8pt}
\label{tab:realworld}
\adjustbox{width=1.0\linewidth}{
\begin{tabular}{cc|ccc}
\toprule
\multicolumn{2}{c|}{Task Suite} & Cucumber Plate & Cube Plate & Cube Bowl \\ \midrule
\multicolumn{1}{c|}{\multirow{5}{*}{\begin{tabular}[c]{@{}c@{}}Static \\ Oracle\end{tabular}}} & $e=5$ & 91.5 $\pm$ 12.7 & 0.0 $\pm$ 0.0  & 76.0 $\pm$ 40.7  \\
\multicolumn{1}{c|}{}  & $e=10$ & 94.0 $\pm$ 11.5 & 81.5 $\pm$ 35.4 & 97.5 $\pm$ 4.2  \\
\multicolumn{1}{c|}{}  & $e=20$ & 88.0 $\pm$ 20.3 & 54.0 $\pm$ 43.9 & 89.5 $\pm$ 17.6 \\
\multicolumn{1}{c|}{} & $e=30$ & 89.0 $\pm$ 19.0 & 74.5 $\pm$ 36.3 & 87.5 $\pm$ 17.0 \\
\multicolumn{1}{c|}{} & $e=40$ & 79.0 $\pm$ 24.5 & 81.5 $\pm$ 28.7 & 56.5 $\pm$ 31.3 \\
\multicolumn{1}{c|}{}  & $e=50$ & 50.0 $\pm$ 40.8 & 51.5 $\pm$ 38.6 & 58.5 $\pm$ 34.0 \\ \midrule
\multicolumn{2}{c|}{Static Oracle+} & 97.0 $\pm$ 7.9 & 81.5 $\pm$ 35.4 & 97.5 $\pm$ 4.2\\ \midrule
\multicolumn{2}{c|}{Random} & 88.5 $\pm$ 14.7 & 60.5 $\pm$ 44.1 & 77.0 $\pm$ 27.3 \\ \midrule
\multicolumn{2}{c|}{\textbf{AutoHorizon}} & \textbf{98.0 $\pm$ 4.8} & \textbf{92.0 $\pm$ 15.7} & \textbf{99.0 $\pm$ 2.1} \\ \bottomrule
\end{tabular}
}
\vspace{-12pt}
\end{table}

\noindent\textbf{Results.}
Tab.~\ref{tab:realworld} summarizes the performance across the three tasks.
The occasionally large standard deviations arise from the binary nature of the outcomes, where some rollouts successfully complete the task, while others fail entirely.
Several key observations emerge from the execution process.
When the execution horizon is too short (\ie, $e \in [1, 5]$), the robot frequently hesitates or stalls during motion.
This behavior stems from the policy’s tendency to predict subtle, low-amplitude movements for the initial few actions within a chunk, resulting in insufficient overall progression.
At moderate horizons ($e \in [20, 40]$), the robot often overreaches or collides with the workspace, reflecting a loss of reactivity.
When the execution horizon becomes excessively long ($e > 40$), the robot struggles to maintain accurate object localization, leading to frequent object drops and a diminished ability to correct errors during execution.

In contrast, AutoHorizon dynamically adjusts the execution horizon throughout the rollout.
As shown in Fig.~\ref{fig:overview}, under stable conditions—such as reaching for or transporting the object—the estimated horizons increase, accelerating execution progress.
When the robot begins to physically interact with the environment (\eg, grasping or placing the cube), the estimated horizons adaptively shorten, enhancing reactivity to environmental changes. 



%% file: sec/5_conclusion.tex
\section{Conclusion}
\label{sec:conclusion}
Determining the execution horizon in action chunking flow policies remains an underexplored yet crucial challenge. 
In this work, we analyze the predictive behaviors of flow-based VLAs through attention weight inspection.
Our analysis reveals that predicted action chunks exhibit limited temporal adaptability and consistently rely on radial action sinks for structural guidance.
Building on these insights, we interpret action self-attention weights as implicit indicators of the model’s predictive confidence and propose an autonomous, attention-guided execution horizon estimation algorithm that dynamically assigns chunk-specific horizons.
Extensive evaluations in both simulated and real-world robotic manipulation tasks demonstrate that our method consistently outperforms other baselines, highlighting its effectiveness and generalizability.

%% file: sec/X_suppl.tex
\clearpage
\setcounter{page}{1}
\maketitlesupplementary

\section{Proof of Proposition 1}
\label{sec:proof}
Under the assumptions of Proposition~1, let $L$ be divisible by $e$ (for simplicity), and let
$\delta^{d}_{j}(e) = k\,e\log e$ with $k>0$ (independent of $j$). 
If $m=L/e$ denotes the number of executed chunks, then there are $m-1$ chunk transitions. 
Hence, from Eq.~\eqref{eq:total_loss} the total loss can be written as
\begin{align}
\mathcal{L}(e) 
&= (m-1)\,\delta^{c} + \sum_{j=1}^{m} \delta^{d}_{j}(e) \notag \\
&= \Big(\frac{L}{e}-1\Big)\delta^{c} + \frac{L}{e}\,k\,e\log e \notag \\
&= \Big(\frac{L}{e}-1\Big)\delta^{c} + Lk\log e. 
\end{align}
Treating $e>0$ as a continuous variable, differentiate:
\begin{align}
\frac{\partial \mathcal{L}(e)}{\partial e} 
&= -\,\frac{L\delta^{c}}{e^{2}} + \frac{Lk}{e}
= \frac{L}{e^{2}}\,\big(k e - \delta^{c}\big). 
\end{align}
The unique stationary point is at $\hat e = \delta^{c}/k$. 
Moreover,
\begin{align}
\frac{\partial^{2} \mathcal{L}(e)}{\partial e^{2}} 
&= \frac{L}{e^{3}}\big(2\delta^{c} - k e\big),
\end{align}
\begin{align}
\frac{\partial^{2} \mathcal{L}(\hat e)}{\partial e^{2}} 
= \frac{L\,\delta^{c}}{\hat e^{3}} \;>\; 0,
\end{align}
so $\mathcal{L}(e)$ is strictly decreasing on $(0,\hat e)$ and strictly increasing on $(\hat e,\infty)$, hence $\hat e$ is the unique global minimizer in the continuous domain. 
(Thus $\mathcal{L}$ is \emph{unimodal}; it need not be globally convex.)

Since $e\in\mathbb{N}$ (and, in practice, $1\le e\le p$), the discrete minimizer lies among the two integers nearest to $\hat e$:
\begin{equation}
\label{eq:ediscrete}
e^{*} \in 
\arg\min_{\,e \in \{\lfloor \hat e \rfloor,\, \lceil \hat e \rceil\}\cap[1,p]}
\mathcal{L}(e),
\qquad
\hat e = \frac{\delta^{c}}{k}.
\end{equation}
Equivalently, without the upper bound $p$,
\begin{equation}
e^{*} \in 
\Big\{
\max\!\big(1,\,\lceil \hat e \rceil\big),\;
\max\!\big(1,\,\lceil \hat e \rceil - 1\big)
\Big\},
\end{equation}
choosing the value that yields the smaller $\mathcal{L}(e)$. 
Finally, since $\lim_{e\to\infty}\mathcal{L}(e)=+\infty$ and $\mathcal{L}$ is unimodal, this discrete minimizer is global.

\section{Method Demonstration}
\begin{figure}[h]
    \centering
    \includegraphics[width=1.0\linewidth, trim=300 10 380 10, clip]{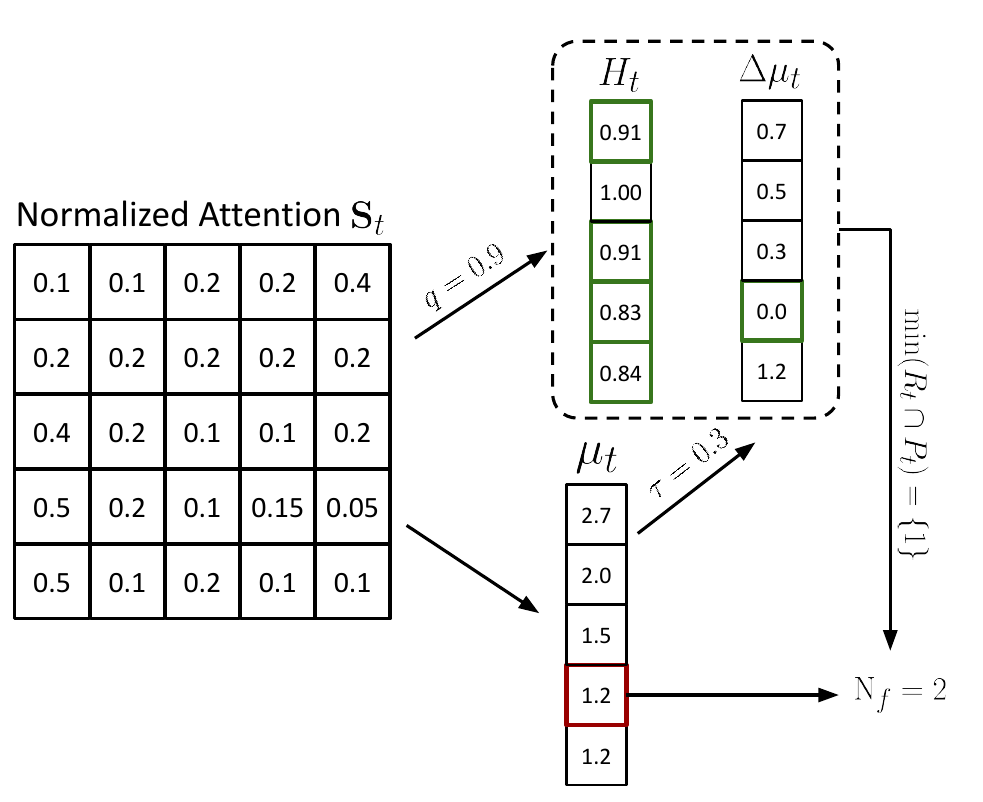}
    \vspace{-16pt}
    \caption{\textbf{Example demonstration of how AutoHorizon works.}}
    \label{fig:method_example}
    \vspace{-6pt}
\end{figure}

We provide an algorithm demonstration in Alg.~\ref{alg:autohorizon}, detailing the proposed method. Also, Fig.~\ref{fig:method_example} illustrates AutoHorizon on a toy example rollout, clarifying the steps for estimating the per-chunk execution horizon: row-wise normalization, low-entropy row selection, forward soft-pointer plateau detection (the backward pointer is not shown), and horizon fusion.

\begin{algorithm}[t]
\caption{AutoHorizon}
\label{alg:autohorizon}
\begin{algorithmic}[1]
\Require
Attention matrix $\mathbf{S}_t$ and its reverse $\tilde{\mathbf{S}}_t$ at time $t$, quantile $q$, threshold $\tau$
\State Obtain row entropy $H_t$ with Eq. (5)
\State Retain low-entropy rows with Eq. (4)

\noindent{\textcolor{lightblue}{// \textbf{Forward traversal}}}
\State Compute $\mu_t$ with Eq. (6)
\State $\Delta \mu_t \gets \texttt{diff}(\mu_t)$
\State $P_t \gets \{\, i \mid \Delta\mu_t[i] < \tau \,\}$
\State $N_f \gets \left\lfloor \mu_t\!\left[\min(R_t \cap P_t)\right] \right\rfloor + 1$

\noindent{\textcolor{lightblue}{// \textbf{Backward traversal}}}
\State Apply step 3-6 to $\tilde{\mathbf{S}}_t$
\State Obtain backward horizon $N_b$

\noindent{\textcolor{lightblue}{// \textbf{Horizon determination}}} \\
\noindent \textbf{if} $N_f + N_b \ge p$ \textbf{then} $N \gets p$ \textbf{else} $N \gets N_f$

\Ensure $N$
\end{algorithmic}
\end{algorithm}

\section{Additional Results}
\label{sec:additional_res}
\subsection{More Baselines}
We further include two re-planning baselines for comparison, including:
\begin{itemize}
    \item \textbf{Action Trigger:} Based on the action chunk distribution, it sets the execution horizon as the first index where the difference between consecutive actions exceeds $\tau_a$.
    \item \textbf{Uncertainty Proxy:} It samples 4 chunks per observation and estimates per-step uncertainty via Monte Carlo variance, calculating the horizon as the earliest step where uncertainty exceeds the threshold of $\tau_u$. The executed actions are obtained by averaging sampled chunks.
\end{itemize}
Tab.~\ref{tab:more_baselines} shows the results. AutoHorizon consistently outperforms these two baselines over various threshold settings. We also observe that these two baselines are very sensitive to the change of hyper-parameters. 
Moreover, Uncertainty Proxy incurs much more computation latency as it needs to compute 4 chunks for every observation, hindering its applicability under real-world settings.  

\begin{table}[h]
\vspace{-6pt}
    \centering
    \caption{Comparison with additional baselines.}
    \vspace{-10pt}
    \adjustbox{width=1.0\linewidth}{
    \begin{tabular}{cc|ccccc}
    \toprule
        \multicolumn{2}{c|}{Task Suite}  & LIB-Spatial & LIB-Object & LIB-Goal & LIB-10 & Press Stapler \\ \midrule
        \multicolumn{1}{c|}{} & $\tau_a=1e^{-2}$ & 95.2$\pm$0.3 & 90.5$\pm$1.3 & 84.9$\pm$2.4 & 77.7$\pm$1.1 & 32.0$\pm$0.0 \\
        \multicolumn{1}{c|}{\multirow{2}{*}{\begin{tabular}[c]{@{}c@{}}Action \\ Trigger\end{tabular}}} & $\tau_a=5e^{-2}$ & 94.4$\pm$0.9 & 94.3$\pm$1.3 & 85.3$\pm$1.5 & 77.5$\pm$0.9 & 58.0$\pm$0.0 \\
        \multicolumn{1}{c|}{} & $\tau_a=1e^{-1}$ & 93.9$\pm$1.9 & 88.9$\pm$2.6 & 85.7$\pm$0.9 & 79.6$\pm$0.7 & 60.0$\pm$0.0 \\
        \multicolumn{1}{c|}{} & $\tau_a=3e^{-1}$ & 79.1$\pm$0.1 & 74.9$\pm$1.9 & 76.9$\pm$0.8 & 74.8$\pm$0.9 & 71.0$\pm$0.0 \\ \midrule
        \multicolumn{1}{c|}{} & $\tau_u=1e^{-2}$ & 94.9$\pm$1.1 & 84.5$\pm$0.7 & 84.4$\pm$0.9 & 70.3$\pm$1.1 & 40.0$\pm$0.0 \\
        \multicolumn{1}{c|}{\multirow{2}{*}{\begin{tabular}[c]{@{}c@{}}Uncertainty \\ Proxy\end{tabular}}} & $\tau_u=5e^{-2}$ & 94.8$\pm$0.7 & 84.8$\pm$1.6 & 85.7$\pm$3.8 & 68.9$\pm$2.8 & 72.0$\pm$0.0 \\
        \multicolumn{1}{c|}{} & $\tau_u=1e^{-1}$ & 94.9$\pm$0.2 & 89.5$\pm$0.8 & 87.7$\pm$1.6 & 79.2$\pm$1.1 & 71.0$\pm$0.0 \\
        \multicolumn{1}{c|}{} & $\tau_u=3e^{-1}$ & 91.5$\pm$0.5 & 82.0$\pm$2.6 & 87.2$\pm$1.1 & 77.6$\pm$2.0 & 72.0$\pm$0.0 \\ \midrule
        \multicolumn{2}{c|}{\textbf{Ours}}  & \textbf{96.5$\pm$0.9} & \textbf{98.0$\pm$0.6} & \textbf{94.4$\pm$1.0} & \textbf{92.1$\pm$1.0} & \textbf{75.0$\pm$0.0} \\
    \bottomrule
    \end{tabular}
    }
    \label{tab:more_baselines}
    \vspace{-10pt}
\end{table}

\subsection{Execution Horizons for Static Oracle+}
\label{sec:static_plus}
Table~\ref{tab:oracle_plus} lists the fixed execution horizons selected for \textit{Static Oracle+} on the LIBERO benchmark. 
Note that we do not sweep the entire range of possible values; instead, we perform a targeted grid search over a plausible interval where the optimal static $e$ is likely to lie.

\begin{table}[h]
\centering
\small
\caption{\textbf{Execution horizons found by Static Oracle+.}}
\vspace{-8pt}
\adjustbox{width=1.0\linewidth}{
\begin{tabular}{c|cccc}
\toprule
Task Suite & LIB-Spatial & LIB-Object & LIB-Goal & LIB-10 \\ \midrule
$\pi_{0.5}$ $(p=10)$ & 10 & 8 & 10 & 10 \\
$\pi_{0.5}$ $(p=50)$ & 5 & 13 & 15 & 15 \\
GR00T $(p=16)$ & 2 & 8 & 8 & 16 \\ \bottomrule
\end{tabular}
}
\label{tab:oracle_plus}
\vspace{-6pt}
\end{table}

The results show substantial variability across tasks and models: the optimal static execution horizon differs markedly, and no clear, generalizable pattern emerges from final success rates alone. 
For example, \textit{LIBERO-Spatial} tends to prefer shorter execution horizons as the prediction horizon increases, whereas \textit{LIBERO-10} exhibits the opposite trend.

\subsection{Comparison with Nearest Static Oracle}
To assess the benefit of the dynamic horizon strategy, we compare AutoHorizon with \textit{Static Oracle} baselines whose fixed horizons are the nearest neighbors to the mean horizon estimated by our method, such that: $e \in \{\left\lfloor e^* \right\rfloor, \left\lfloor e^* \right\rfloor + 1 \}$. 
Table~\ref{tab:near_oracle} reports results on the LIBERO benchmark using $\pi_{0.5}$ with $p=50$, and the corresponding mean horizons from our method are shown in Fig.~\ref{fig:boxplot}.

\begin{table}[h]
\centering
\small
\caption{\textbf{Comparison with Nearest Static Oracle on LIBERO.} $m$ indicates the mean of the estimated execution horizon distribution by our method.}
\vspace{-8pt}
\adjustbox{width=1.0\linewidth}{
\begin{tabular}{c|cccc}
\toprule
Task Suite & LIB-Spatial & LIB-Object & LIB-Goal & LIB-10 \\ \midrule
Static Oracle ($e=\left \lfloor m\right \rfloor$) & 95.7 $\pm$ 0.2 & 97.6 $\pm$ 0.6 & 92.4 $\pm$ 0.0 & 89.1 $\pm$ 1.4 \\
Static Oracle ($e=\left \lceil m\right \rceil$) & 95.1 $\pm$ 0.2 & 97.2 $\pm$ 0.0 & 93.9 $\pm$ 0.2 & 91.9 $\pm$ 0.4 \\
\textbf{Ours} & \textbf{96.5 $\pm$ 0.9} & \textbf{98.0 $\pm$ 0.6} & \textbf{94.4 $\pm$ 1.0} & \textbf{92.1 $\pm$ 1.0} \\ \bottomrule
\end{tabular}
}
\vspace{-12pt}
\label{tab:near_oracle}
\end{table}

Across different task configurations, AutoHorizon achieves higher or comparable success rates relative to the nearest \textit{Static Oracle}, demonstrating the effectiveness of dynamic horizon estimation. 
We further observe that the mean execution horizon produced by AutoHorizon closely approximates the optimal static choice, while its occasional selection of larger or smaller horizons enables handling corner cases during long rollouts—an advantage unattainable with a single fixed horizon.

\subsection{Performance under Shorter Prediction Horizon}
\begin{figure}[h]
    \vspace{-6pt}
    \centering
    \includegraphics[width=1.0\linewidth, trim=10 10 10 10, clip]{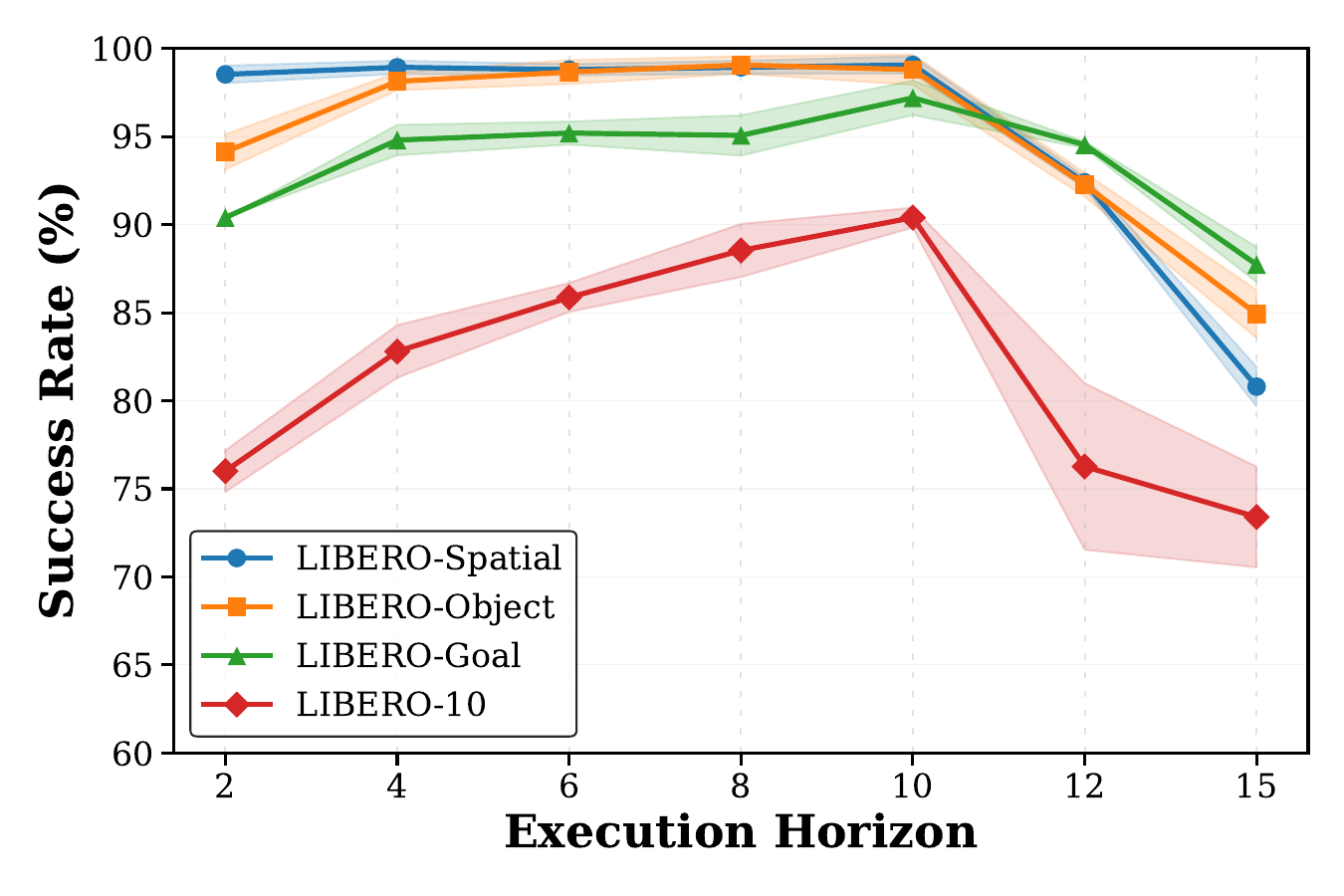}
    \vspace{-24pt}
    \caption{\textbf{Average success rates on the LIBERO benchmark with a prediction horizon of 10 using $\pi_{0.5}$.}}
    \label{fig:supp_h10}
    \vspace{-12pt}
\end{figure}
Fig.~\ref{fig:supp_h10} reports results under a shorter prediction horizon ($p = 10$) using $\pi_{0.5}$. 
The relationship between performance and execution-horizon persists, with peak performance often attained at the boundary $e = p$. 
When $e > p$, performance drops sharply. 
We attribute this to a train–test mismatch: the policy is trained on chunks of length at most $p$ and does not generalize to longer horizons. 
This observation provides a natural upper bound for the execution horizon such that $1 \le e \le p$.

\begin{figure*}[t]
    \centering
    \includegraphics[width=1.0\linewidth, trim=10 120 10 80, clip]{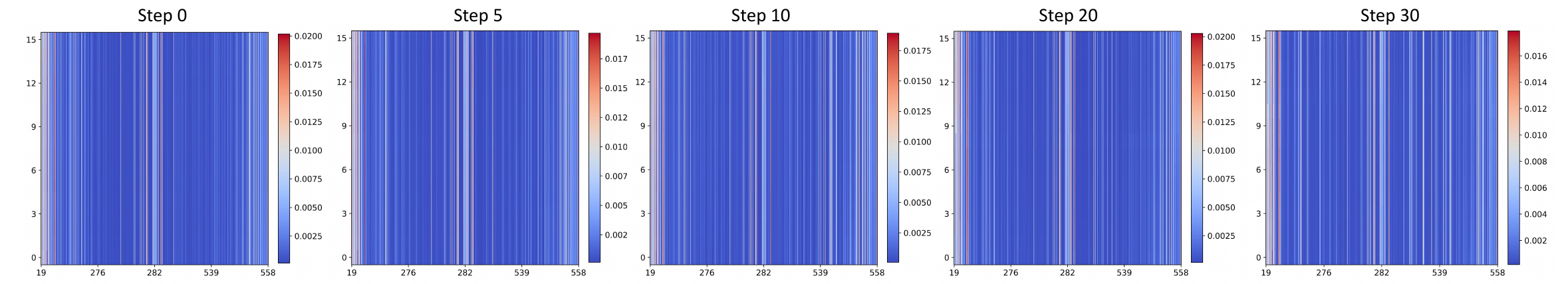}
    \vspace{-18pt}
    \caption{\textbf{Visualization of the cross-attention weights in GR00T~N1.5 over different rollout steps.}}
    \label{fig:gr00t_cross}
    \vspace{-6pt}
\end{figure*}

\subsection{Effect of Language Tokens}
\label{sec:lang_effect}
We investigate the role of language tokens in action generation by analyzing their contribution within the cross-attention mechanism of $\pi_{0.5}$ on the LIBERO benchmark. Although the language tokens exhibit high attention weights, we find that they carry only subtle semantic influence on action prediction.
To quantify this effect, we perform ablation studies summarized in Tab.~\ref{tab:lang_effect}.
\textit{Original} denotes using the unaltered attention weights, while \textit{Mask Lang} indicates fully masking the language tokens’ attention weights (with re-normalization applied to preserve the softmax distribution).

The results show that masking the language tokens leads to mixed but generally minor performance changes, often resulting in only marginal degradation.
This suggests that most linguistic information has been absorbed into the vision tokens during training, rendering explicit language attention largely redundant at inference time.

To further examine this redundancy, we apply the Visual Attention Redistribution (VAR) technique proposed by \citet{visual_attnsink}, which redistributes a portion of the attention mass from certain tokens to others. Under our case, we redistribute the attention mass from the language tokens to other modalities.
We denote this variant as \textit{VAR-L-$p$}, where $p$ represents the fraction of language attention weights redistributed (\eg, \textit{VAR-L-0.5} redistributes 50\% of the language attention).
This strategy acts as a soft version of language masking, since \textit{Original} corresponds to \textit{VAR-L-1.0}.
When setting $p = 0.5$, performance notably improves—sometimes even surpassing the \textit{Original} baseline (\eg, cases under $p=50$, $e=10$).
This finding indicates that partially redistributing language attention to vision–action tokens can enhance model performance, reinforcing the conclusion that language token attention contains substantial redundancy in pretrained VLAs.

\begin{table}[h]
\centering
\small
\vspace{-6pt}
\caption{\textbf{Effect of language tokens on LIBERO benchmark.}}
\vspace{-8pt}
\adjustbox{width=1.0\linewidth}{
\begin{tabular}{cc|cccc}
\toprule
\multicolumn{2}{c|}{Task Suite} & LIB-Spatial & LIB-Object & LIB-Goal & LIB-10 \\ \midrule
\multicolumn{1}{c|}{\multirow{3}{*}{\begin{tabular}[c]{@{}c@{}} $p=10$ \\ $e=10$\end{tabular}}} & Original & 99.1 $\pm$ 0.5  & 98.8 $\pm$ 0.9  & 97.2 $\pm$ 1.0 & 90.4 $\pm$ 0.6 \\
\multicolumn{1}{c|}{}  & Mask Lang & 97.7 $\pm$ 0.5 & 99.5 $\pm$ 0.5 &  96.1 $\pm$ 0.8 & 91.3 $\pm$ 1.6 \\
\multicolumn{1}{c|}{} & VAR-L-0.5 & 98.5 $\pm$ 0.5 & 99.2 $\pm$ 0.6 & 96.9 $\pm$ 1.0 & 90.7 $\pm$ 1.0  \\ \midrule
\multicolumn{1}{c|}{\multirow{3}{*}{\begin{tabular}[c]{@{}c@{}} $p=50$ \\ $e=10$\end{tabular}}} & Original & 91.2 $\pm$ 2.3 & 94.9 $\pm$ 0.2 & 97.1 $\pm$ 0.2 & 92.7 $\pm$ 1.0 \\
\multicolumn{1}{c|}{}  & Mask Lang & 92.8 $\pm$ 0.9 & 93.7 $\pm$ 0.5 & 81.7 $\pm$ 1.5 & 79.6 $\pm$ 1.5 \\
\multicolumn{1}{c|}{} & VAR-L-0.5 & 94.8 $\pm$ 1.2 & 96.7 $\pm$ 0.8 & 89.6 $\pm$ 1.6 & 92.1 $\pm$ 1.5 \\ 
\bottomrule
\end{tabular}
}
\label{tab:lang_effect}
\vspace{-12pt}
\end{table}

\subsection{Smoothness Analysis}
\begin{figure}[h]
\vspace{-6pt}
    \includegraphics[width=0.95\linewidth, trim=0 0 20 0, clip]{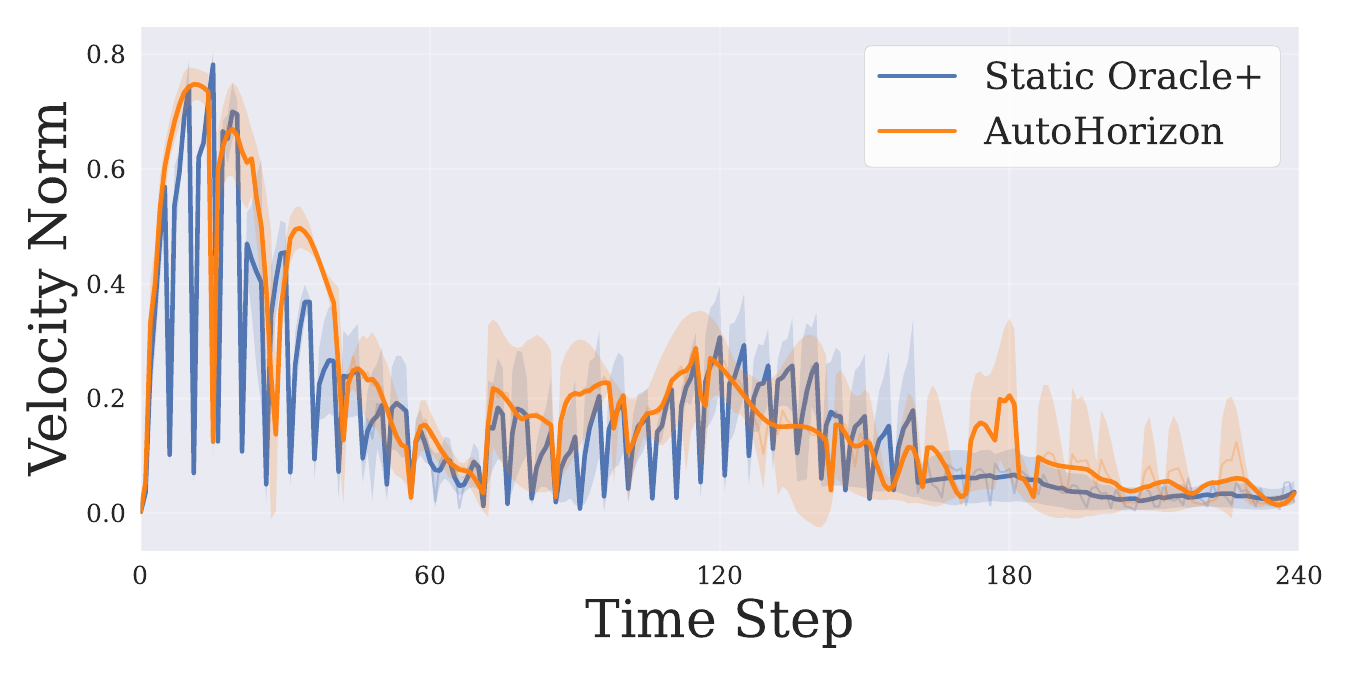}
    \vspace{-12pt}
    \captionof{figure}{Smoothness analysis.}
    \label{fig:velocity_norms}
    \vspace{-12pt}
\end{figure}

We visualize the robot kinematics over time through calculating the joint velocity norm. Fig.~\ref{fig:velocity_norms} shows that AutoHorizon produces smoother velocity transitions over time, demonstrating its potential of achieving stable and smooth actions with improved performance.

\subsection{Hyper-parameter Sensitivity}
\label{sec:hyper_abl}
To verify the effect of hyper-parameter choices, we conducted experiments using $\pi_{0.5}$ on LIBERO-10. Tab.~\ref{tab:sensitivity} shows that AutoHorizon remains stable across variations in attention choice $L$, entropy quantile $q$, and threshold $\tau$. During analysis, we fix the other parameters choices for fair comparison. We also provide a brief analysis on the effect of combining different hyper-parameters in Tab.~\ref{tab:hyp_abl}, where the model performance remains stable across a wide range of settings.

\begin{table}[h]
    \centering
    \small
    \vspace{-6pt}
    \caption{\textbf{Hyper-parameter sensitivity analysis.}}
    \vspace{-8pt}
    \adjustbox{width=0.8\linewidth}{
    \begin{tabular}{c|c|c|c|c}
    \toprule
    $L = 2$ & $L = 3$ & $L = 4$ & $L = 5$ & $L = 6$ \\ 
    89.9$\pm$1.2 & 92.1$\pm$1.0 & 90.9$\pm$2.5 & 89.5$\pm$1.8 & 90.4$\pm$1.4 \\ \midrule
     $q = 0.7$ & $q = 0.8$ & $q = 0.9$ & $q = 0.99$ & $q = 0.999$ \\
    90.8$\pm$0.3 & 92.0$\pm$1.4 & 92.1$\pm$1.0 & 91.9$\pm$1.2 & 91.5$\pm$2.0 \\ \midrule
     $\tau = 0.1$ & $\tau = 0.2$ & $\tau = 0.3$ & $\tau = 0.4$ & $\tau = 0.5$ \\ 
    90.9$\pm$1.0 & 90.0$\pm$0.9 & 92.1$\pm$1.0 & 91.7$\pm$1.0 & 91.6$\pm$1.4 \\ \bottomrule
    \end{tabular}
    }
    \label{tab:sensitivity}
    \vspace{-8pt}
\end{table}

\begin{table}[h]
\centering
\small
\vspace{-6pt}
\caption{\textbf{More ablations on hyper-parameters.}}
\vspace{-8pt}
\adjustbox{width=0.8\linewidth}{
\begin{tabular}{c|cccc}
\toprule
Params & $q=0.8$ & $q=0.9$ & $q=0.99$ \\ \midrule
$\tau=0.2$ & 90.0 $\pm$ 0.9 & 90.5 $\pm$ 1.9 & 90.0 $\pm$ 1.7 \\ 
$\tau=0.3$ & 92.0 $\pm$ 1.4 & 92.1 $\pm$ 1.0 & 91.9 $\pm$ 1.2 \\
$\tau=0.4$ & 91.7 $\pm$ 1.0 & 88.3 $\pm$ 0.8 & 90.0 $\pm$ 2.1 \\
\bottomrule
\end{tabular}
}
\vspace{-12pt}
\label{tab:hyp_abl}
\end{table}

\section{Attention Mechanism within GR00T~N1.5}
\label{sec:gr00t}
We further demonstrate that our observations are not limited to $\pi_{0.5}$ but also generalize to other flow-based VLA models such as GR00T~N1.5, which employs a substantially different architecture and training pipeline.
Unlike $\pi_{0.5}$, GR00T~N1.5 adopts an alternating modality fusion process: one block performs cross-attention between action and vision–language tokens, followed by another block that applies action self-attention.
This alternating pattern repeats multiple times during the forward pass, enabling iterative refinement of multimodal representations.

\begin{figure}[h]
    \centering
    \includegraphics[width=1.0\linewidth, trim=340 120 340 75, clip]{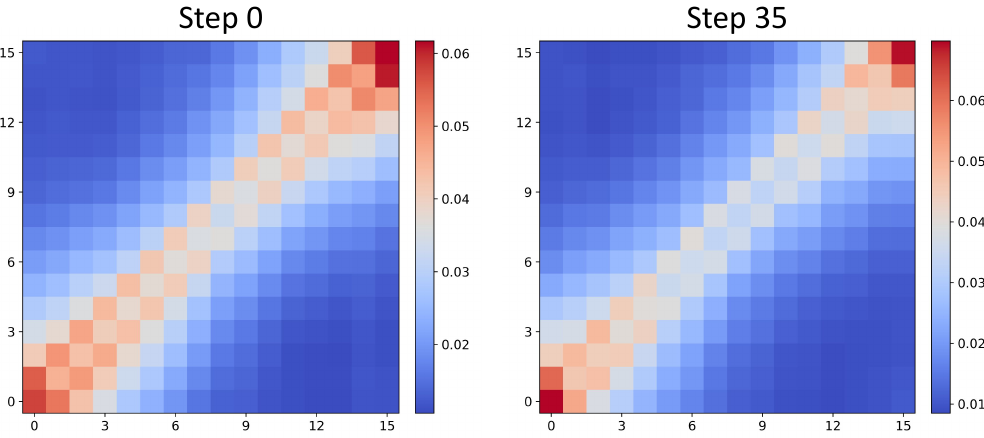}
    \vspace{-18pt}
    \caption{\textbf{Visualization of normalized action self-attention in GR00T~N1.5 over different rollout steps.} }
    \label{fig:gr00t_action}
    \vspace{-6pt}
\end{figure}

In Fig.~\ref{fig:gr00t_cross}, we visualize the cross-attention maps between generated actions and vision–language tokens over different policy execution steps.
The token indices $0$–$19$, $276$–$282$, and $539$ to the end correspond to visual input.
The visualization reveals that the cross-attention mechanism in GR00T~N1.5 exhibits the same invariance pattern observed in $\pi_{0.5}$—different generated actions consistently attend to the same vision–language tokens.

In Fig.~\ref{fig:gr00t_action}, we visualize the action self-attention maps over different rollout steps.
GR00T~N1.5 likewise displays the radial action sink phenomenon, where the initial and terminal action tokens receive disproportionately high attention weights.
This consistent structural bias across models reinforces our interpretation that radial action sinks act as implicit indicators of the model’s predictive limits and can therefore inform execution horizon estimation.


%% file: main.bib
@String(CVPR= {IEEE Conf. Comput. Vis. Pattern Recog.})

@String(ICLR = {Int. Conf. Learn. Represent.})

@String(CVPR  = {CVPR})

@String(ICLR  = {ICLR})

@article{vla_survey1,
  author={Kento Kawaharazuka and Jihoon Oh and Jun Yamada and Ingmar Posner and Yuke Zhu},
  journal={IEEE Access},
  title={Vision-Language-Action Models for Robotics: A Review Towards Real-World Applications},
  volume={13},
  pages={162467--162504},
  year={2025},
  doi={10.1109/ACCESS.2025.3609980},
}

@misc{vla_survey2,
      title={A Survey on Vision-Language-Action Models for Embodied AI}, 
      author={Yueen Ma and Zixing Song and Yuzheng Zhuang and Jianye Hao and Irwin King},
      year={2025},
      eprint={2405.14093},
      archivePrefix={arXiv},
      primaryClass={cs.RO},
      url={https://arxiv.org/abs/2405.14093}, 
}

@misc{pi05,
      title={$\pi_{0.5}$: a Vision-Language-Action Model with Open-World Generalization}, 
      author={Physical Intelligence and Kevin Black and Noah Brown and James Darpinian and Karan Dhabalia and Danny Driess and Adnan Esmail and Michael Equi and Chelsea Finn and Niccolo Fusai and Manuel Y. Galliker and Dibya Ghosh and Lachy Groom and Karol Hausman and Brian Ichter and Szymon Jakubczak and Tim Jones and Liyiming Ke and Devin LeBlanc and Sergey Levine and Adrian Li-Bell and Mohith Mothukuri and Suraj Nair and Karl Pertsch and Allen Z. Ren and Lucy Xiaoyang Shi and Laura Smith and Jost Tobias Springenberg and Kyle Stachowicz and James Tanner and Quan Vuong and Homer Walke and Anna Walling and Haohuan Wang and Lili Yu and Ury Zhilinsky},
      year={2025},
      eprint={2504.16054},
      archivePrefix={arXiv},
      primaryClass={cs.LG},
      url={https://arxiv.org/abs/2504.16054}, 
}

@misc{pi0,
      title={$\pi_0$: A Vision-Language-Action Flow Model for General Robot Control}, 
      author={Kevin Black and Noah Brown and Danny Driess and Adnan Esmail and Michael Equi and Chelsea Finn and Niccolo Fusai and Lachy Groom and Karol Hausman and Brian Ichter and Szymon Jakubczak and Tim Jones and Liyiming Ke and Sergey Levine and Adrian Li-Bell and Mohith Mothukuri and Suraj Nair and Karl Pertsch and Lucy Xiaoyang Shi and James Tanner and Quan Vuong and Anna Walling and Haohuan Wang and Ury Zhilinsky},
      year={2024},
      eprint={2410.24164},
      archivePrefix={arXiv},
      primaryClass={cs.LG},
      url={https://arxiv.org/abs/2410.24164}, 
}

@misc{openvla,
      title={OpenVLA: An Open-Source Vision-Language-Action Model}, 
      author={Moo Jin Kim and Karl Pertsch and Siddharth Karamcheti and Ted Xiao and Ashwin Balakrishna and Suraj Nair and Rafael Rafailov and Ethan Foster and Grace Lam and Pannag Sanketi and Quan Vuong and Thomas Kollar and Benjamin Burchfiel and Russ Tedrake and Dorsa Sadigh and Sergey Levine and Percy Liang and Chelsea Finn},
      year={2024},
      eprint={2406.09246},
      archivePrefix={arXiv},
      primaryClass={cs.RO},
      url={https://arxiv.org/abs/2406.09246}, 
}

@misc{smolvla,
      title={SmolVLA: A Vision-Language-Action Model for Affordable and Efficient Robotics}, 
      author={Mustafa Shukor and Dana Aubakirova and Francesco Capuano and Pepijn Kooijmans and Steven Palma and Adil Zouitine and Michel Aractingi and Caroline Pascal and Martino Russi and Andres Marafioti and Simon Alibert and Matthieu Cord and Thomas Wolf and Remi Cadene},
      year={2025},
      eprint={2506.01844},
      archivePrefix={arXiv},
      primaryClass={cs.LG},
      url={https://arxiv.org/abs/2506.01844}, 
}

@misc{dp,
      title={Diffusion Policy: Visuomotor Policy Learning via Action Diffusion}, 
      author={Cheng Chi and Zhenjia Xu and Siyuan Feng and Eric Cousineau and Yilun Du and Benjamin Burchfiel and Russ Tedrake and Shuran Song},
      year={2024},
      eprint={2303.04137},
      archivePrefix={arXiv},
      primaryClass={cs.RO},
      url={https://arxiv.org/abs/2303.04137}, 
}

@misc{dp3,
      title={3D Diffusion Policy: Generalizable Visuomotor Policy Learning via Simple 3D Representations}, 
      author={Yanjie Ze and Gu Zhang and Kangning Zhang and Chenyuan Hu and Muhan Wang and Huazhe Xu},
      year={2024},
      eprint={2403.03954},
      archivePrefix={arXiv},
      primaryClass={cs.RO},
      url={https://arxiv.org/abs/2403.03954}, 
}

@misc{act,
      title={Learning Fine-Grained Bimanual Manipulation with Low-Cost Hardware}, 
      author={Tony Z. Zhao and Vikash Kumar and Sergey Levine and Chelsea Finn},
      year={2023},
      eprint={2304.13705},
      archivePrefix={arXiv},
      primaryClass={cs.RO},
      url={https://arxiv.org/abs/2304.13705}, 
}

@article{bid,
    title={Bidirectional Decoding: Improving Action Chunking via Closed-Loop Resampling},
    author={Liu, Yuejiang and Hamid, Jubayer Ibn and Xie, Annie and Lee, Yoonho and Du, Max and Finn, Chelsea},
    year={2025},
    journal={International Conference on Learning Representations (ICLR)},
    url={https://arxiv.org/abs/2408.17355}
}

@misc{rtc,
      title={Real-Time Execution of Action Chunking Flow Policies}, 
      author={Kevin Black and Manuel Y. Galliker and Sergey Levine},
      year={2025},
      eprint={2506.07339},
      archivePrefix={arXiv},
      primaryClass={cs.RO},
      url={https://arxiv.org/abs/2506.07339}, 
}

@misc{paligemma,
      title={PaliGemma: A versatile 3B VLM for transfer}, 
      author={Lucas Beyer and Andreas Steiner and André Susano Pinto and Alexander Kolesnikov and Xiao Wang and Daniel Salz and Maxim Neumann and Ibrahim Alabdulmohsin and Michael Tschannen and Emanuele Bugliarello and Thomas Unterthiner and Daniel Keysers and Skanda Koppula and Fangyu Liu and Adam Grycner and Alexey Gritsenko and Neil Houlsby and Manoj Kumar and Keran Rong and Julian Eisenschlos and Rishabh Kabra and Matthias Bauer and Matko Bošnjak and Xi Chen and Matthias Minderer and Paul Voigtlaender and Ioana Bica and Ivana Balazevic and Joan Puigcerver and Pinelopi Papalampidi and Olivier Henaff and Xi Xiong and Radu Soricut and Jeremiah Harmsen and Xiaohua Zhai},
      year={2024},
      eprint={2407.07726},
      archivePrefix={arXiv},
      primaryClass={cs.CV},
      url={https://arxiv.org/abs/2407.07726}, 
}

@misc{qwen3,
      title={Qwen3 Technical Report}, 
      author={An Yang and Anfeng Li and Baosong Yang and Beichen Zhang and Binyuan Hui and Bo Zheng and Bowen Yu and Chang Gao and Chengen Huang and Chenxu Lv and Chujie Zheng and Dayiheng Liu and Fan Zhou and Fei Huang and Feng Hu and Hao Ge and Haoran Wei and Huan Lin and Jialong Tang and Jian Yang and Jianhong Tu and Jianwei Zhang and Jianxin Yang and Jiaxi Yang and Jing Zhou and Jingren Zhou and Junyang Lin and Kai Dang and Keqin Bao and Kexin Yang and Le Yu and Lianghao Deng and Mei Li and Mingfeng Xue and Mingze Li and Pei Zhang and Peng Wang and Qin Zhu and Rui Men and Ruize Gao and Shixuan Liu and Shuang Luo and Tianhao Li and Tianyi Tang and Wenbiao Yin and Xingzhang Ren and Xinyu Wang and Xinyu Zhang and Xuancheng Ren and Yang Fan and Yang Su and Yichang Zhang and Yinger Zhang and Yu Wan and Yuqiong Liu and Zekun Wang and Zeyu Cui and Zhenru Zhang and Zhipeng Zhou and Zihan Qiu},
      year={2025},
      eprint={2505.09388},
      archivePrefix={arXiv},
      primaryClass={cs.CL},
      url={https://arxiv.org/abs/2505.09388}, 
}

@misc{gpt4,
      title={GPT-4 Technical Report}, 
      author={OpenAI},
      year={2024},
      eprint={2303.08774},
      archivePrefix={arXiv},
      primaryClass={cs.CL},
      url={https://arxiv.org/abs/2303.08774}, 
}

@misc{gemini,
      title={Gemini: A Family of Highly Capable Multimodal Models}, 
      author={Gemini Team},
      year={2025},
      eprint={2312.11805},
      archivePrefix={arXiv},
      primaryClass={cs.CL},
      url={https://arxiv.org/abs/2312.11805}, 
}

@misc{eagle25,
      title={Eagle 2.5: Boosting Long-Context Post-Training for Frontier Vision-Language Models}, 
      author={Guo Chen and Zhiqi Li and Shihao Wang and Jindong Jiang and Yicheng Liu and Lidong Lu and De-An Huang and Wonmin Byeon and Matthieu Le and Tuomas Rintamaki and Tyler Poon and Max Ehrlich and Tuomas Rintamaki and Tyler Poon and Tong Lu and Limin Wang and Bryan Catanzaro and Jan Kautz and Andrew Tao and Zhiding Yu and Guilin Liu},
      year={2025},
      eprint={2504.15271},
      archivePrefix={arXiv},
      primaryClass={cs.CV},
      url={https://arxiv.org/abs/2504.15271}, 
}

@misc{internvl3,
      title={InternVL3: Exploring Advanced Training and Test-Time Recipes for Open-Source Multimodal Models}, 
      author={Jinguo Zhu and Weiyun Wang and Zhe Chen and Zhaoyang Liu and Shenglong Ye and Lixin Gu and Hao Tian and Yuchen Duan and Weijie Su and Jie Shao and Zhangwei Gao and Erfei Cui and Xuehui Wang and Yue Cao and Yangzhou Liu and Xingguang Wei and Hongjie Zhang and Haomin Wang and Weiye Xu and Hao Li and Jiahao Wang and Nianchen Deng and Songze Li and Yinan He and Tan Jiang and Jiapeng Luo and Yi Wang and Conghui He and Botian Shi and Xingcheng Zhang and Wenqi Shao and Junjun He and Yingtong Xiong and Wenwen Qu and Peng Sun and Penglong Jiao and Han Lv and Lijun Wu and Kaipeng Zhang and Huipeng Deng and Jiaye Ge and Kai Chen and Limin Wang and Min Dou and Lewei Lu and Xizhou Zhu and Tong Lu and Dahua Lin and Yu Qiao and Jifeng Dai and Wenhai Wang},
      year={2025},
      eprint={2504.10479},
      archivePrefix={arXiv},
      primaryClass={cs.CV},
      url={https://arxiv.org/abs/2504.10479}, 
}

@InProceedings{rt2,
  title = 	 {RT-2: Vision-Language-Action Models Transfer Web Knowledge to Robotic Control},
  author =       {Zitkovich, Brianna and Yu, Tianhe and Xu, Sichun and Xu, Peng and Xiao, Ted and Xia, Fei and Wu, Jialin and Wohlhart, Paul and Welker, Stefan and Wahid, Ayzaan and Vuong, Quan and Vanhoucke, Vincent and Tran, Huong and Soricut, Radu and Singh, Anikait and Singh, Jaspiar and Sermanet, Pierre and Sanketi, Pannag R. and Salazar, Grecia and Ryoo, Michael S. and Reymann, Krista and Rao, Kanishka and Pertsch, Karl and Mordatch, Igor and Michalewski, Henryk and Lu, Yao and Levine, Sergey and Lee, Lisa and Lee, Tsang-Wei Edward and Leal, Isabel and Kuang, Yuheng and Kalashnikov, Dmitry and Julian, Ryan and Joshi, Nikhil J. and Irpan, Alex and Ichter, Brian and Hsu, Jasmine and Herzog, Alexander and Hausman, Karol and Gopalakrishnan, Keerthana and Fu, Chuyuan and Florence, Pete and Finn, Chelsea and Dubey, Kumar Avinava and Driess, Danny and Ding, Tianli and Choromanski, Krzysztof Marcin and Chen, Xi and Chebotar, Yevgen and Carbajal, Justice and Brown, Noah and Brohan, Anthony and Arenas, Montserrat Gonzalez and Han, Kehang},
  booktitle = 	 {Proceedings of The 7th Conference on Robot Learning},
  pages = 	 {2165--2183},
  year = 	 {2023},
  editor = 	 {Tan, Jie and Toussaint, Marc and Darvish, Kourosh},
  volume = 	 {229},
  series = 	 {Proceedings of Machine Learning Research},
  month = 	 {06--09 Nov},
  publisher =    {PMLR},
  url = 	 {https://proceedings.mlr.press/v229/zitkovich23a.html},
}

@inproceedings{cotvla,
  title={Cot-vla: Visual chain-of-thought reasoning for vision-language-action models},
  author={Zhao, Qingqing and Lu, Yao and Kim, Moo Jin and Fu, Zipeng and Zhang, Zhuoyang and Wu, Yecheng and Li, Zhaoshuo and Ma, Qianli and Han, Song and Finn, Chelsea and others},
  booktitle={Proceedings of the Computer Vision and Pattern Recognition Conference},
  pages={1702--1713},
  year={2025}
}

@article{tinyvla,
  title={Tinyvla: Towards fast, data-efficient vision-language-action models for robotic manipulation},
  author={Wen, Junjie and Zhu, Yichen and Li, Jinming and Zhu, Minjie and Tang, Zhibin and Wu, Kun and Xu, Zhiyuan and Liu, Ning and Cheng, Ran and Shen, Chaomin and others},
  journal={IEEE Robotics and Automation Letters},
  year={2025},
  publisher={IEEE}
}

@article{dreamvla,
  title={Dreamvla: a vision-language-action model dreamed with comprehensive world knowledge},
  author={Zhang, Wenyao and Liu, Hongsi and Qi, Zekun and Wang, Yunnan and Yu, Xinqiang and Zhang, Jiazhao and Dong, Runpei and He, Jiawei and Lu, Fan and Wang, He and others},
  journal={arXiv preprint arXiv:2507.04447},
  year={2025}
}

@article{pointvla,
  title={Pointvla: Injecting the 3d world into vision-language-action models},
  author={Li, Chengmeng and Wen, Junjie and Peng, Yan and Peng, Yaxin and Feng, Feifei and Zhu, Yichen},
  journal={arXiv preprint arXiv:2503.07511},
  year={2025}
}

@article{3dvla,
  title={3d-vla: A 3d vision-language-action generative world model},
  author={Zhen, Haoyu and Qiu, Xiaowen and Chen, Peihao and Yang, Jincheng and Yan, Xin and Du, Yilun and Hong, Yining and Gan, Chuang},
  journal={arXiv preprint arXiv:2403.09631},
  year={2024}
}

@misc{gr00t,
      title={GR00T N1: An Open Foundation Model for Generalist Humanoid Robots}, 
      author={NVIDIA and : and Johan Bjorck and Fernando Castañeda and Nikita Cherniadev and Xingye Da and Runyu Ding and Linxi "Jim" Fan and Yu Fang and Dieter Fox and Fengyuan Hu and Spencer Huang and Joel Jang and Zhenyu Jiang and Jan Kautz and Kaushil Kundalia and Lawrence Lao and Zhiqi Li and Zongyu Lin and Kevin Lin and Guilin Liu and Edith Llontop and Loic Magne and Ajay Mandlekar and Avnish Narayan and Soroush Nasiriany and Scott Reed and You Liang Tan and Guanzhi Wang and Zu Wang and Jing Wang and Qi Wang and Jiannan Xiang and Yuqi Xie and Yinzhen Xu and Zhenjia Xu and Seonghyeon Ye and Zhiding Yu and Ao Zhang and Hao Zhang and Yizhou Zhao and Ruijie Zheng and Yuke Zhu},
      year={2025},
      eprint={2503.14734},
      archivePrefix={arXiv},
      primaryClass={cs.RO},
      url={https://arxiv.org/abs/2503.14734}, 
}

@misc{streamingllm,
      title={Efficient Streaming Language Models with Attention Sinks}, 
      author={Guangxuan Xiao and Yuandong Tian and Beidi Chen and Song Han and Mike Lewis},
      year={2024},
      eprint={2309.17453},
      archivePrefix={arXiv},
      primaryClass={cs.CL},
      url={https://arxiv.org/abs/2309.17453}, 
}

@misc{visual_attnsink,
      title={See What You Are Told: Visual Attention Sink in Large Multimodal Models}, 
      author={Seil Kang and Jinyeong Kim and Junhyeok Kim and Seong Jae Hwang},
      year={2025},
      eprint={2503.03321},
      archivePrefix={arXiv},
      primaryClass={cs.CV},
      url={https://arxiv.org/abs/2503.03321}, 
}

@misc{h2o,
      title={H$_2$O: Heavy-Hitter Oracle for Efficient Generative Inference of Large Language Models}, 
      author={Zhenyu Zhang and Ying Sheng and Tianyi Zhou and Tianlong Chen and Lianmin Zheng and Ruisi Cai and Zhao Song and Yuandong Tian and Christopher Ré and Clark Barrett and Zhangyang Wang and Beidi Chen},
      year={2023},
      eprint={2306.14048},
      archivePrefix={arXiv},
      primaryClass={cs.LG},
      url={https://arxiv.org/abs/2306.14048}, 
}

@article{sink_llm,
  title={When attention sink emerges in language models: An empirical view},
  author={Gu, Xiangming and Pang, Tianyu and Du, Chao and Liu, Qian and Zhang, Fengzhuo and Du, Cunxiao and Wang, Ye and Lin, Min},
  journal={arXiv preprint arXiv:2410.10781},
  year={2024}
}

@misc{libero,
      title={LIBERO: Benchmarking Knowledge Transfer for Lifelong Robot Learning}, 
      author={Bo Liu and Yifeng Zhu and Chongkai Gao and Yihao Feng and Qiang Liu and Yuke Zhu and Peter Stone},
      year={2023},
      eprint={2306.03310},
      archivePrefix={arXiv},
      primaryClass={cs.AI},
      url={https://arxiv.org/abs/2306.03310}, 
}

@InProceedings{robotwin,
    author    = {Mu, Yao and Chen, Tianxing and Chen, Zanxin and Peng, Shijia and Lan, Zhiqian and Gao, Zeyu and Liang, Zhixuan and Yu, Qiaojun and Zou, Yude and Xu, Mingkun and Lin, Lunkai and Xie, Zhiqiang and Ding, Mingyu and Luo, Ping},
    title     = {RoboTwin: Dual-Arm Robot Benchmark with Generative Digital Twins},
    booktitle = {Proceedings of the Computer Vision and Pattern Recognition Conference (CVPR)},
    month     = {June},
    year      = {2025},
    pages     = {27649-27660}
}

@article{robotwin2.0,
  title={Robotwin 2.0: A scalable data generator and benchmark with strong domain randomization for robust bimanual robotic manipulation},
  author={Chen, Tianxing and Chen, Zanxin and Chen, Baijun and Cai, Zijian and Liu, Yibin and Li, Zixuan and Liang, Qiwei and Lin, Xianliang and Ge, Yiheng and Gu, Zhenyu and others},
  journal={arXiv preprint arXiv:2506.18088},
  year={2025}
}

@ARTICLE{franka,
  author={Haddadin, Sami},
  journal={IEEE Robotics \& Automation Magazine}, 
  title={The Franka Emika Robot: A Standard Platform in Robotics Research}, 
  year={2024},
  volume={31},
  number={4},
  pages={136-148},
  doi={10.1109/MRA.2024.3451788}
}

@misc{droid,
      title={DROID: A Large-Scale In-The-Wild Robot Manipulation Dataset}, 
      author={Alexander Khazatsky and Karl Pertsch and Suraj Nair and Ashwin Balakrishna and Sudeep Dasari and Siddharth Karamcheti and Soroush Nasiriany and Mohan Kumar Srirama and Lawrence Yunliang Chen and Kirsty Ellis and Peter David Fagan and Joey Hejna and Masha Itkina and Marion Lepert and Yecheng Jason Ma and Patrick Tree Miller and Jimmy Wu and Suneel Belkhale and Shivin Dass and Huy Ha and Arhan Jain and Abraham Lee and Youngwoon Lee and Marius Memmel and Sungjae Park and Ilija Radosavovic and Kaiyuan Wang and Albert Zhan and Kevin Black and Cheng Chi and Kyle Beltran Hatch and Shan Lin and Jingpei Lu and Jean Mercat and Abdul Rehman and Pannag R Sanketi and Archit Sharma and Cody Simpson and Quan Vuong and Homer Rich Walke and Blake Wulfe and Ted Xiao and Jonathan Heewon Yang and Arefeh Yavary and Tony Z. Zhao and Christopher Agia and Rohan Baijal and Mateo Guaman Castro and Daphne Chen and Qiuyu Chen and Trinity Chung and Jaimyn Drake and Ethan Paul Foster and Jensen Gao and Vitor Guizilini and David Antonio Herrera and Minho Heo and Kyle Hsu and Jiaheng Hu and Muhammad Zubair Irshad and Donovon Jackson and Charlotte Le and Yunshuang Li and Kevin Lin and Roy Lin and Zehan Ma and Abhiram Maddukuri and Suvir Mirchandani and Daniel Morton and Tony Nguyen and Abigail O'Neill and Rosario Scalise and Derick Seale and Victor Son and Stephen Tian and Emi Tran and Andrew E. Wang and Yilin Wu and Annie Xie and Jingyun Yang and Patrick Yin and Yunchu Zhang and Osbert Bastani and Glen Berseth and Jeannette Bohg and Ken Goldberg and Abhinav Gupta and Abhishek Gupta and Dinesh Jayaraman and Joseph J Lim and Jitendra Malik and Roberto Martín-Martín and Subramanian Ramamoorthy and Dorsa Sadigh and Shuran Song and Jiajun Wu and Michael C. Yip and Yuke Zhu and Thomas Kollar and Sergey Levine and Chelsea Finn},
      year={2025},
      eprint={2403.12945},
      archivePrefix={arXiv},
      primaryClass={cs.RO},
      url={https://arxiv.org/abs/2403.12945}, 
}

@misc{vpp,
      title={Video Prediction Policy: A Generalist Robot Policy with Predictive Visual Representations}, 
      author={Yucheng Hu and Yanjiang Guo and Pengchao Wang and Xiaoyu Chen and Yen-Jen Wang and Jianke Zhang and Koushil Sreenath and Chaochao Lu and Jianyu Chen},
      year={2025},
      eprint={2412.14803},
      archivePrefix={arXiv},
      primaryClass={cs.CV},
      url={https://arxiv.org/abs/2412.14803}, 
}

@article{bc,
    title = {A survey of robot learning from demonstration},
    journal = {Robotics and Autonomous Systems},
    volume = {57},
    number = {5},
    pages = {469-483},
    year = {2009},
    issn = {0921-8890},
    doi = {https://doi.org/10.1016/j.robot.2008.10.024},
    url = {https://www.sciencedirect.com/science/article/pii/S0921889008001772},
    author = {Brenna D. Argall and Sonia Chernova and Manuela Veloso and Brett Browning},
}

@misc{imitation,
      title={A Survey of Imitation Learning: Algorithms, Recent Developments, and Challenges}, 
      author={Maryam Zare and Parham M. Kebria and Abbas Khosravi and Saeid Nahavandi},
      year={2023},
      eprint={2309.02473},
      archivePrefix={arXiv},
      primaryClass={cs.LG},
      url={https://arxiv.org/abs/2309.02473}, 
}

@misc{attention,
      title={Attention Is All You Need}, 
      author={Ashish Vaswani and Noam Shazeer and Niki Parmar and Jakob Uszkoreit and Llion Jones and Aidan N. Gomez and Lukasz Kaiser and Illia Polosukhin},
      year={2023},
      eprint={1706.03762},
      archivePrefix={arXiv},
      primaryClass={cs.CL},
      url={https://arxiv.org/abs/1706.03762}, 
}

@misc{hirobot,
      title={Hi Robot: Open-Ended Instruction Following with Hierarchical Vision-Language-Action Models}, 
      author={Lucy Xiaoyang Shi and Brian Ichter and Michael Equi and Liyiming Ke and Karl Pertsch and Quan Vuong and James Tanner and Anna Walling and Haohuan Wang and Niccolo Fusai and Adrian Li-Bell and Danny Driess and Lachy Groom and Sergey Levine and Chelsea Finn},
      year={2025},
      eprint={2502.19417},
      archivePrefix={arXiv},
      primaryClass={cs.RO},
      url={https://arxiv.org/abs/2502.19417}, 
}

@misc{gr2,
      title={GR-2: A Generative Video-Language-Action Model with Web-Scale Knowledge for Robot Manipulation}, 
      author={Chi-Lam Cheang and Guangzeng Chen and Ya Jing and Tao Kong and Hang Li and Yifeng Li and Yuxiao Liu and Hongtao Wu and Jiafeng Xu and Yichu Yang and Hanbo Zhang and Minzhao Zhu},
      year={2024},
      eprint={2410.06158},
      archivePrefix={arXiv},
      primaryClass={cs.RO},
      url={https://arxiv.org/abs/2410.06158}, 
}

@misc{uva,
      title={Unified Video Action Model}, 
      author={Shuang Li and Yihuai Gao and Dorsa Sadigh and Shuran Song},
      year={2025},
      eprint={2503.00200},
      archivePrefix={arXiv},
      primaryClass={cs.RO},
      url={https://arxiv.org/abs/2503.00200}, 
}

@article{action_chunk,
  title={Action chunking as policy compression},
  author={Lai, Lucy and Huang, Ann Zixiang and Gershman, Samuel J},
  journal={PsyArXiv},
  year={2022}
}

@inproceedings{pigdm,
  title={Pseudoinverse-Guided Diffusion Models for Inverse Problems},
  author={Jiaming Song and Arash Vahdat and Morteza Mardani and Jan Kautz},
  booktitle={International Conference on Learning Representations},
  year={2023},
  url={https://api.semanticscholar.org/CorpusID:259298715}
}
